%% file: paper.tex
\documentclass[letterpaper, 10 pt, conference]{ieeeconf}  % Comment this line out if you need a4paper

\IEEEoverridecommandlockouts                              % This command is only needed if 
\usepackage{times}
\usepackage{epsfig}
\usepackage{graphicx}
\usepackage{amsmath}
\usepackage{amssymb}
\usepackage{booktabs}
\usepackage{soul}
\usepackage{lipsum}  % Holger: to generate dummy text
\usepackage{xcolor} % Holger: To use named colors
\usepackage{multirow}
\usepackage{algorithm}
\usepackage{algpseudocode}
\usepackage{authblk}

\setlipsum{%
  par-before = \begingroup\color{lightgray},
  par-after = \endgroup
}
\usepackage{array}
\newcolumntype{C}[1]{>{\centering\let\newline\\\arraybackslash\hspace{0pt}}m{#1}}
\newcolumntype{L}[1]{>{\raggedright\let\newline\\\arraybackslash\hspace{0pt}}m{#1}}
\newcolumntype{R}[1]{>{\raggedleft\let\newline\\\arraybackslash\hspace{0pt}}m{#1}}

\usepackage{tabularx}

\usepackage{mathtools}
\usepackage[pagebackref=true,breaklinks=true,colorlinks,bookmarks=false]{hyperref}

\title{\LARGE \bf
Towards learning-based planning:\\ The nuPlan benchmark for real-world autonomous driving
}

\author{Napat Karnchanachari$^{*}$
\and
Dimitris Geromichalos$^{*}$
\and
Kok Seang Tan
\and
Nanxiang Li
\and
Christopher Eriksen
\and
Shakiba Yaghoubi 
\and
Noushin Mehdipour 
\and
Gianmarco Bernasconi 
\and
Whye Kit Fong 
\and
Yiluan Guo 
\and
Holger Caesar$^{\dag}$
\thanks{
Authors are or were $(^{\dag})$ with Motional. $(^{*})$ indicates equal contribution.}
}

\begin{document}

\maketitle
\thispagestyle{empty}
\pagestyle{empty}

%%%%%%%%%% ABSTRACT
\begin{abstract}
  \input{sections/abstract.tex}
\end{abstract}

%%%%%%%%% BODY TEXT
\input{sections/introduction.tex}
\input{sections/relatedwork.tex}
\input{sections/dataset.tex}

\input{sections/simulation.tex}
\input{sections/experiments.tex}
\input{sections/conclusion.tex}
\input{sections/acknowledgement.tex}

%%%%%%%%%%%%%%%%%%%%%%%%%%%%%%%%%%%%%%%%%%%%%%%%%%%%%%%%%%%%%%%%%%%%%%%%%%%%%%%%
\bibliographystyle{IEEEtran}
\bibliography{paper}

\newpage
\clearpage
%\appendix
\section*{\large{\textbf{Supplementary Material}}}
\input{sections/appendix.tex}

\end{document}

%% file: sections/abstract.tex
% Introduction
Machine Learning (ML) has replaced handcrafted methods for perception and prediction in autonomous vehicles. 
Yet for the equally important planning task, the adoption of ML-based techniques is slow.
We present nuPlan, the world's first real-world autonomous driving dataset and benchmark.
% Goal
The benchmark is designed to test the ability of ML-based planners to handle diverse driving situations and to make safe and efficient decisions.
% Overview
We introduce a new large-scale dataset that consists of 1282 hours of diverse driving scenarios from 4 cities (Las Vegas, Boston, Pittsburgh, and Singapore) and includes high-quality auto-labeled object tracks and traffic light data. We mine and taxonomize common \& rare driving scenarios which are used during evaluation to get fine-grained insights into the performance and characteristics of a planner.
Beyond the dataset, we provide a simulation and evaluation framework that enables a planner's actions to be simulated in closed-loop to account for interactions with other traffic participants.
% Baselines
We present a detailed analysis of numerous baselines and investigate gaps between ML-based and traditional methods.
% Code
Find the nuPlan dataset and code at \url{nuplan.org}.

%% file: sections/introduction.tex
\section{Introduction}
% Why are learning-based planners not successful?
\PARstart{I}{n} the last decade, autonomous vehicle perception and prediction have been revolutionized by deep learning-based methods trained on large-scale datasets~\cite{zhou2018voxelnet,pointpillars,vora2020pointpainting,chen2021polarstream,centerpoint,phan2020covernet,liang2020learning}. 
While similar attempts have been made in the field of learning-based or neural planning, these are not yet able to surpass their rule-based counterparts.
One possible reason is the difficulty of generalizing driving scenarios when learned from a limited number of examples.
Furthermore, driving scenarios typically follow a long-tail distribution, which further exacerbates the generalization issue.
Finally, learning-based planning lacks formal safety guarantees, thus making it potentially unsafe and challenging to certify.

% Role of nuPlan
We introduce the nuPlan dataset and simulation framework for autonomous vehicle planning.
Our goal is to create a testbed for open-loop and closed-loop planning starting in real-world scenarios.
This test bed is then used to compare traditional, learning-based, and hybrid planners.
nuPlan enables numerous novel types of research, such as learning-based planning, the interplay between prediction and planning, and end-to-end planning using a large amount of published sensor data.
\begin{figure}
  \centering
  \includegraphics[width=0.9\linewidth]
  {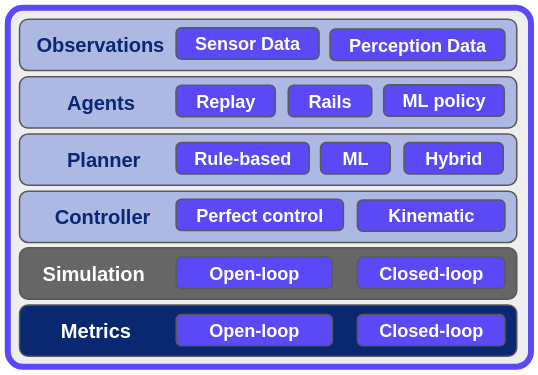}
  \vspace{-2mm}
  \caption{An overview of the nuPlan simulation framework.}
  \label{fig:planning_framework}
  \vspace{-10pt}
\end{figure}
%
% Contributions
We make the following contributions:
\begin{itemize}
    \item We release the largest dataset for autonomous driving to date, with a total of 1282h from 4 cities. We also publish an unprecedented 128h of sensor data.
    \item We develop techniques to auto-label the dataset with accurate object tracks, traffic lights, and scenario labels.
    \item We publish our closed-loop simulation and evaluation framework (Fig.~\ref{fig:planning_framework}) and compare the performance of traditional and learning-based planners to identify gaps.
\end{itemize}

%% file: sections/relatedwork.tex
\section{Related Work}

In this section we discuss the most important datasets and simulators for autonomous vehicle planning. Based on these, we introduce the main categories of planners: classical, learning-based, and hybrid planners.

% - Virtual: Has the advantage that we can control the environment, focus on rare events etc.
% - Real-world: More realistic, capture real OOD events, but requires large-scale mining for them, which we do.

%Lyft~\cite{lyft} was notably larger with 1118h from a single route. While the map was more detailed, the on-car perception is likewise of low quality and the single route limits generalization to other routes.
%Waymo~\cite{waymo} drastically expanded the coverage to 6 cities with a total of 570h of data.
%Most notably they developed an offline perception system~\cite{autolabeling} that is able to produce high quality tracks.
%Despite the enormous scale and comprehensiveness of these datasets, none of them focuses on the planning task.

%The prediction or motion forecasting task is separate from the planning task. It aims to estimate the future trajectories of other agents in the scene, in absence of a known goal of these agents.
%However, the datasets from this task can often be used for planning purposes as well.
\subsection{Datasets}
\label{secsec:datasets}
Tab.~\ref{tab:datasets} provides an overview of large-scale prediction and planning datasets with more than 20h of data.
We omit smaller datasets like Interaction~\cite{interaction}, highD~\cite{high_d}, inD~\cite{in_d}, OpenDD~\cite{open_dd} and CommonRoad~\cite{commonroad} and subsets of datasets not focused on prediction or planning~\cite{argoverse,argoverse2,shifts}.
With the exception of nuPlan and CommonRoad~\cite{commonroad}, all datasets focus on prediction (motion forecasting).
Offline perception~\cite{waymo_offboard} is crucial to train and evaluate planners using high-quality object tracks, but only present in Waymo~\cite{waymo}, MONA~\cite{mona} and nuPlan.
Likewise, the availability of traffic light statuses is crucial for realistic traffic simulation, but only Waymo~\cite{waymo} and Lyft~\cite{lyft} contain these and only from an online perception system, rather than developing offline traffic light status inference as in nuPlan.
To assess planning performance we need to focus on specific scenarios and evaluate them in closed-loop. 
nuPlan is the first dataset to feature both scenario tags and a closed-loop simulation framework.
Lyft~\cite{lyft} provides interactive tutorials for closed-loop simulation, but they lack the modular framework, evaluation server, and hold-out test set of nuPlan.
Finally, we need a large-scale dataset to generalize well. 
Only the Lyft~\cite{lyft}, Shifts~\cite{shifts} and nuPlan datasets provide more than 1000h of driving data and nuPlan was the first such dataset that provides lidar and camera sensor data (128h), although Waymo~\cite{waymo} later also released compressed lidar data. 

% Decision making dataset survey https://arxiv.org/pdf/2306.16784.pdf
\begin{table*}[t] 
  \vspace{4pt}
  \centering
  \caption{An overview of datasets for prediction and planning (* see Sec.~\ref{secsec:datasets} for a detailed discussion)}
  \vspace{-4mm}
  %\resizebox{1.02\textwidth}{!}{
  \begin{tabularx}{\textwidth}{ c|C{6mm}C{14mm}C{14mm}ccccc}
   \bottomrule
   Dataset      & Year      & Tasks  &  Perception & Traffic Lights & Scenario Tags & Closed-loop & Total Volume (h) & Sensor Volume (h) \\
   \hline
   %CommonRoad~\cite{commonroad} &   2017    &   Plan.    &   no      & no    &   no  & \textbf{yes} & 0 & 0\\ % handcrafted/simulated scenarios
   Argoverse 1~\cite{argoverse} &   2019    &   Pred    &   online     &   no      &   no  &   no  & 320 & 0\\
   Waymo~\cite{waymo}           &   2019    &   Pred    &\textbf{offline} &   online      &   no  &   no  & 574 & \ 0$^{*}$\\ % waymo added lidar for the past 1s (and not the future 8s) in v.1.2
   Lyft~\cite{lyft}             &   2020    &   Pred    &   online  &   online      &   no  &  \ no$^{*}$  & 1118 & 0\\ % the dataset comes with example notebooks how to do closed loop, but that there is no official evaluation server.
   Shifts~\cite{shifts}         &   2021    &   Pred    &   n/a       &   no       &   no   &   no   & \ \textbf{1667*} & 0\\
   MONA~\cite{mona}             &   2022    &   Pred    & \textbf{offline} &   no      &   no  &   no  & 130 & 0\\
   Argoverse 2~\cite{argoverse2}&   2023    &   Pred    &   online  &   no      &   no  &   no  & 763 & 0\\
   \hline
   nuPlan                       &   2021    & \textbf{Pred+Plan} & \textbf{offline}  & \textbf{offline} & \textbf{yes} & \textbf{yes} & \textbf{1282} & \textbf{128}\\
  \toprule
  \end{tabularx}
  %}
  \vspace{-6mm}
  \label{tab:datasets}
\end{table*}
\subsection{Simulation}
Many works in the literature use proprietary  simulators~\cite{scheel2022urban,ChauffeurNet}.
While sharing their approaches to planning, they do not provide enough information to reproduce their experimental results.
Graphical simulators like CARLA~\cite{carla} and AirSim~\cite{shah2018airsim} focus on photorealistic rendering, but lack the realism of real-world maps and agent behavior.
CommonRoad~\cite{commonroad} was the first open-source simulator focused on planning. 
However, CommonRoad~\cite{commonroad} does not provide a real world dataset as the basis for the simulation, instead resorting to a small number of manually crafted scenarios and tools to import other datasets, albeit without any sensor data.
With nuPlan we aim to overcome the above limitations by releasing a large-scale real-world dataset and an open-source closed-loop simulator.
Following the release of nuPlan, ScenarioNet~\cite{daniels2018scenarionet} focused specifically on Reinforcement Learning and integrated nuPlan and other datasets into their pipeline. 
They also interfaced with MetaDrive~\cite{li2022metadrive} that enables graphical simulation of nuPlan.
% Nocturne [47] with 500+ hours Waymo motion data is merely designed for multi-agent Reinforcement Learning (RL)
% SimNet built on 1000 hours L5 motion dataset targets at scenario generation. Researchers who want to use nuPlan data for multi-agent RL have to spend a significant effort on building a new simulator or bridging Nocturne [47] and nuPlan dataset. 
% VISTA~\cite{amini2022vista} warps image/lidar to new viewpoints in closed-loop
%\holger{
%TODO: Cite paper that does closed-loop on nuScenes \url{https://arxiv.org/pdf/2305.10430.pdf}
%}

\subsection{Planning}
%\holger{Cut planning subsection by 30-50 percent.}

\textbf{Classical planning.}
The planning problem has long been treated as an optimization problem in the traditional approaches~\cite{paden2016survey, buehler2009darpa, ajanovic2018search, fraichard1993path, fan2018baidu, montemerlo2008junior}. By carefully designing a cost function, the optimization aims to generate the optimal trajectory that minimizes the cost function in the corresponding search space (e.g., A* search~\cite{buehler2009darpa, ajanovic2018search}, sampling-based methods~\cite{kuffner2000rrt,karaman2011sampling}, dynamic programming~\cite{fan2018baidu}). While these approaches enjoy the theoretical guarantees on the convergence to an optimal solution, hand-crafting the cost function that represents the human-like driving behavior is challenging. In practice, many studies rely on tremendous engineering efforts to fine-tune the solution.

\textbf{Learning based planning.}
Pioneering by the study of \cite{pomerleau1988alvinn}, the idea of using a neural network to imitate expert driver and directly output driving control command provides an alternative planning solution. With the recent success of deep learning, the learning based planning received considerable attention~\cite{codevilla2019exploring, bansal2018chauffeurnet, bojarski2016end, zeng2019end, hawke2020urban, vitelli2022safetynet, scheel2022urban, hu2023planning}. 
%Learning based planning aims to directly outputting driving command avoids the requirement of hand-crafting the cost function. 
%It also has the potential to scale with data. As more and more expert driving data being collected daily, learning based planning could potentially leverage the increasing training data to improve its performance and handle a wide variety of driving scenarios. 
%However, learning based planning usually lacks of safety guarantees, which prevent it to be deployed at scale in the real world.

\textit{Imitation learning (IL) and inverse reinforcement learning (IRL):}
IL trains a model to either map the sensor data directly (end-to-end system), or indirectly through the perception and prediction models (modularized system), to the expert driver actions (e.g. steering and speed profile)~\cite{pomerleau1988alvinn, bojarski2016end, bansal2018chauffeurnet, hallgarten2023prediction}. 
With the advancements in deep learning literature, IL studies adopt the state-of-the-art supervised learning models architectures to learn better scene representations~\cite{muller2018driving, scheel2022urban}. IL often suffers from a poor generalization where the compounding error leads to driving scenarios that are outside of the training data, known as ``covariate shift"~\cite{pomerleau1988alvinn}. Carefully designed data augmentation is often used to address this issue~\cite{bansal2018chauffeurnet}. 
% shows that using data augmentation can address this issue. However, designing the data augmentation requires careful design and thus can be limited in practice. UrbanDriver~\cite{scheel2022urban} uses the compounding error accumulated over multiple steps during the closed-loop rollouts as the data augmentation. However, such data augmentation is induced by the model error. Its effect can be either unstable (when model switch between different error modes) or diminishing (when the model makes little error on the training data). Moreover, the closed-loop training can be computational expensive. 
As an alternative to directly imitating the driver behavior, IRL aims to learn an unknown reward function that explains expert demonstrations~\cite{abbeel2004apprenticeship}. Once learned, such reward function is used to infer the optimal trajectory from a set of pre-defined or generated trajectories~\cite{ho2016generative}. The maximum entropy formulation of IRL~\cite{aghasadeghi2011maximum} has been applied to autonomous driving where the reward function is estimated based on a set of handcrafted features in \cite{wulfmeier2015maximum,ziebart2008maximum,phan2023driving}.

\textit{Reinforcement learning (RL):}
RL learns the optimal driving behavior by interacting with the environment and optimizing a given reward function. RL is well-suited for handling the interaction between the agent and the environment in a sequential decision process. However, due to its learning by ``trial-and-error" search nature, studies in RL rely on the driving simulation to provide the environment~\cite{shalev2016safe, chen2020learning,chen2021learning}. These studies have demonstrated strong performance in simulations. The real-world applications of RL in autonomous driving are also reported in \cite{riedmiller2007learning,kendall2019learning}. 
%While promising, these results are yet to be evaluated in real-world urban driving. The challenges include the gap between the simulation and real-world environment and the difficulty in specifying the appropriate reward function.

\textbf{Hybrid solutions.}
Hybrid solutions are proposed to leverage the advantages of both the classical and learning based planning. Several studies use learning to improve the classical planning algorithm. These include using a learning model to guide the exploration for sampling-based path-planners~\cite{arslan2015machine}, using a learning model to improve the efficiency of sampling-based motion-planners in high dimensional setting~\cite{qureshi2019motion}, applying optimizer to actively rectifies the learning model’s plan to satisfy the safety and comfort requirements~\cite{pulver2021pilot,vitelli2022safetynet}. Other studies leverage the classical planning to generate the trajectory candidates, which are passed to an ML-based model to evaluate~\cite{phan2023driving, zeng2019end,casas2021mp3}.

\textbf{System design.} 
While planning is the ultimate goal of the autonomous driving system, different system designs to assemble the perception and prediction introduce opportunities and challenges to improve the planning performance.  Most autonomous driving systems use a multi-stage pipeline of independent tasks like perception, prediction and planning~\cite{chen2021data,gu2021densetnt,xiong2023cape}.
The hope is that the performance gain on individual tasks translates to a better planning performance.
In contrast, various studies consider multi-task learning (MTL) where they jointly train models to perform perception, prediction and planning simultaneously~\cite{sadat2020perceive, ngiam2022scene,casas2021mp3,zeng2019end}. These works have shown that MTL achieves better data utilization at lower computation cost. 
Recently, some studies leverage the query-based design in transformer architectures to integrate all tasks in a unified framework that is trainable end-to-end~\cite{chen2022learning,mp3, chitta2021neat,hu2022st,hu2023planning,wu2022trajectory}.
Such a framework encourages better spatio-temporal feature learning from the sensor data and directly improves the planning performance.
\vspace{-1mm}

% End-to-end planning paper
% https://openaccess.thecvf.com/content/CVPR2023/papers/Hu_Planning-Oriented_Autonomous_Driving_CVPR_2023_paper.pdf

% \holger{Add new preprint ``From Prediction to Planning With Goal Conditioned Lane Graph Traversals'' has SOTA results on nuPlan! \url{https://arxiv.org/pdf/2302.07753.pdf}}

%% file: sections/dataset.tex
\section{Dataset}
In this section, we describe how we collect the nuPlan dataset. 
We enhance it by auto-labeling the object tracks of other agents and traffic lights, as well as mining for scenarios that are relevant for tracking.

\begin{figure}
  \centering
  \includegraphics[width=\linewidth]{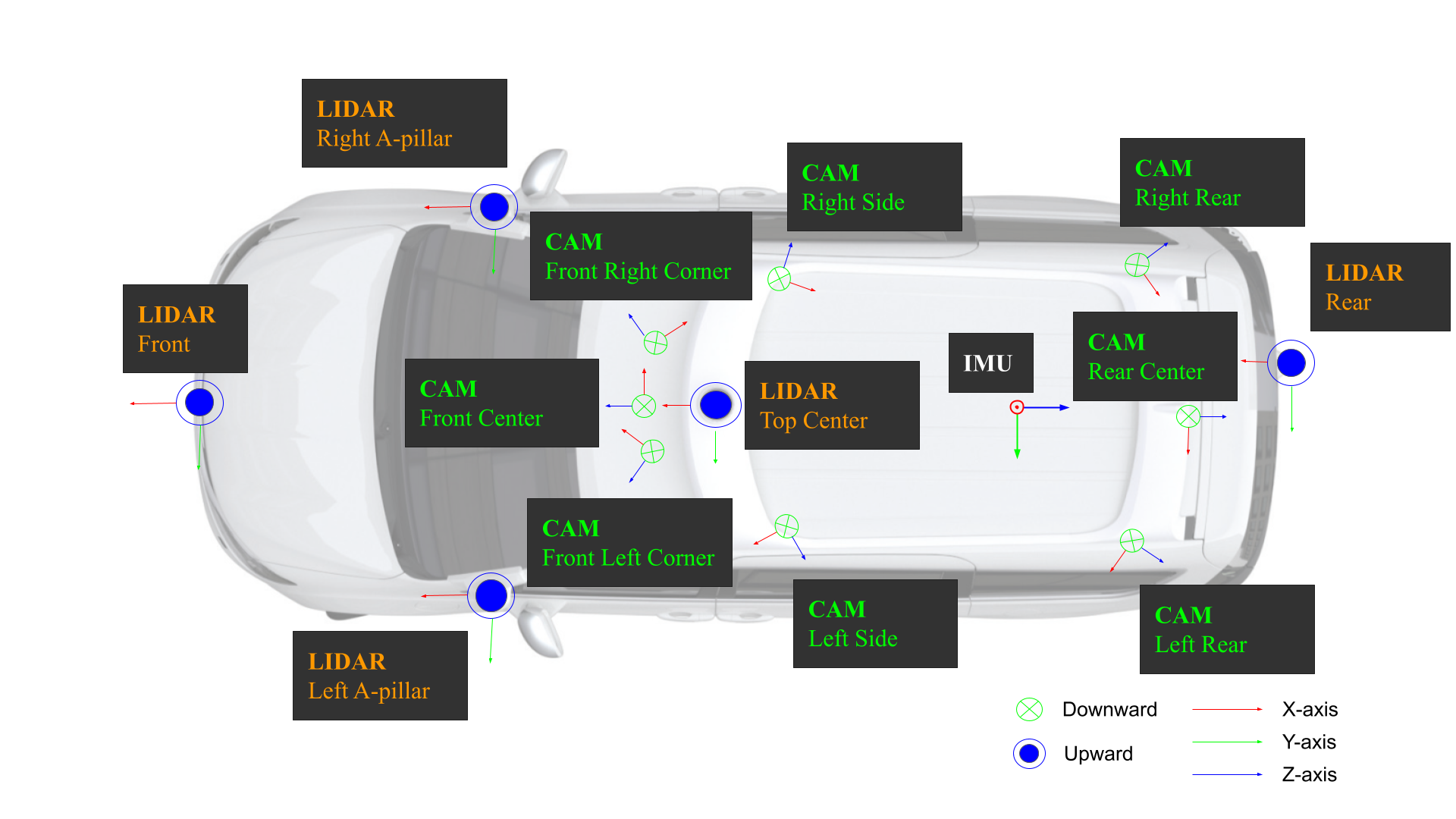}
  \vspace{-8mm}
  \caption{The data collection vehicle's sensor setup.}
  \vspace{-2mm}
  \label{fig:nuplan_car_setup}
\end{figure}

\begin{figure}
  \centering
  \includegraphics[width=0.49\textwidth]{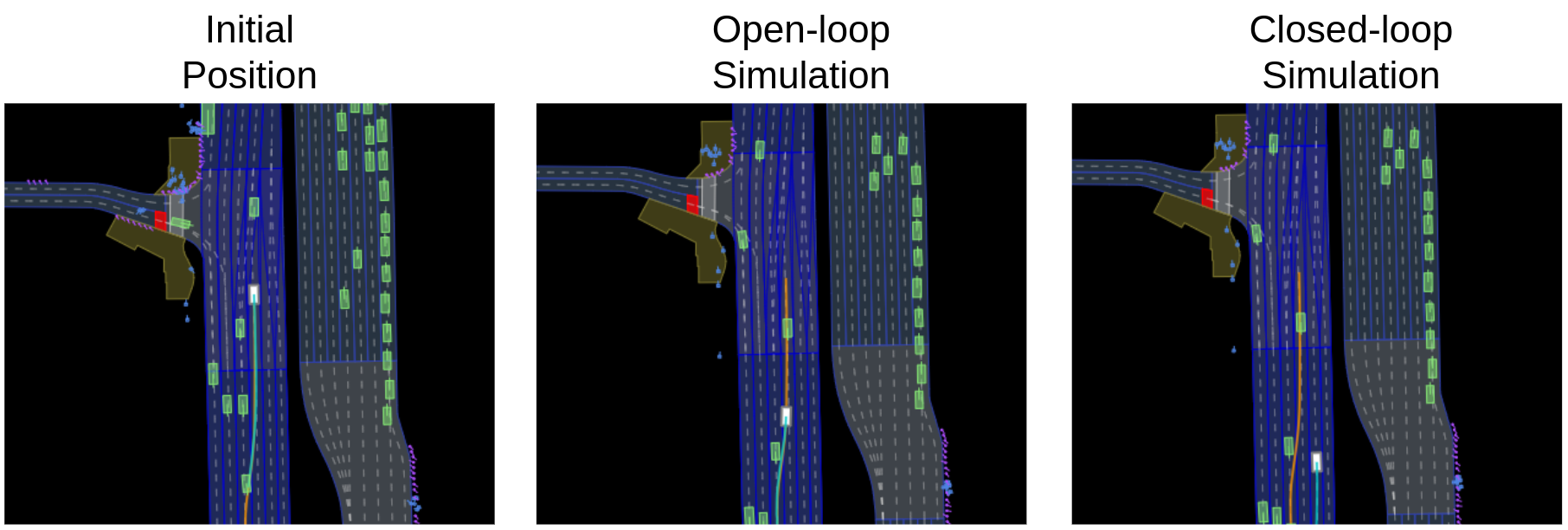}
  \vspace{-8mm}
  \caption{Open-loop (OL) v.s. closed-loop (CL) simulation. In OL, ego vehicle follows the ground-truth trajectories. In CL, ego vehicle follows the planner model output. As shown above, starting from the initial scenario (left), using a trained machine learning planner (Urban Driver model in Sec.~\ref{sec:experiments}), we show the different simulation roll-out in open-loop (middle) and closed-loop (right). [{\color{orange} Ego ground-truth trajectory}, {\color{cyan} Planner output trajectory}]}
  \label{fig:example_nuplan_example}
    \vspace{-10pt}
\end{figure}

\subsection{Data collection}
We collected data from 4 cities (Boston, Pittsburgh, Las Vegas, and Singapore) to build a benchmark dataset for ML-based planning. 
In total, we have 1282 hours of challenging and real-world driving scenarios. 
For example, double parking in Boston, custom precedence patterns for left turns in Pittsburgh, crowded pick-up and drop-off points (PUDOs) in casinos in Las Vegas, and left-hand traffic in Singapore.
We exclude heavy rain and night data, as these would impact the quality of our perception system (see Sec.~\ref{sec:dataset:autolabeling}).

\textbf{Manual driving.} 
We use Chrysler Pacifica Plug-in Hybrid Electric Vehicles (PHEV) to drive in these cities.
See Fig.~\ref{fig:nuplan_car_setup} for the sensor setup. Our vehicle operators (VOs) are instructed to use a natural driving style and drive safely. Since our focus is on planning, it is crucial that we drive manually, while most other datasets~\cite{argoverse,argoverse2,waymo,lyft} use a combination of manual and automated driving, which may lead the planner to imitate less desirable driving behavior.
The VOs drive from a predefined starting point to a goal using a known route. 
For example, we drive between various hotels and casinos on the Las Vegas strip, which are typical routes for our robotaxi and are known and mapped beforehand.

\textbf{Sensor data.} 
Sensor data include lidar point clouds and camera images. 
Due to the vast scale of the full sensor dataset (200+ TB), we only release a subset of the sensor data which totals 128 hours.
This subset was selected to satisfy all stratification constraints as described below.

\begin{figure}
    \centering
    \includegraphics[width=0.85\linewidth]{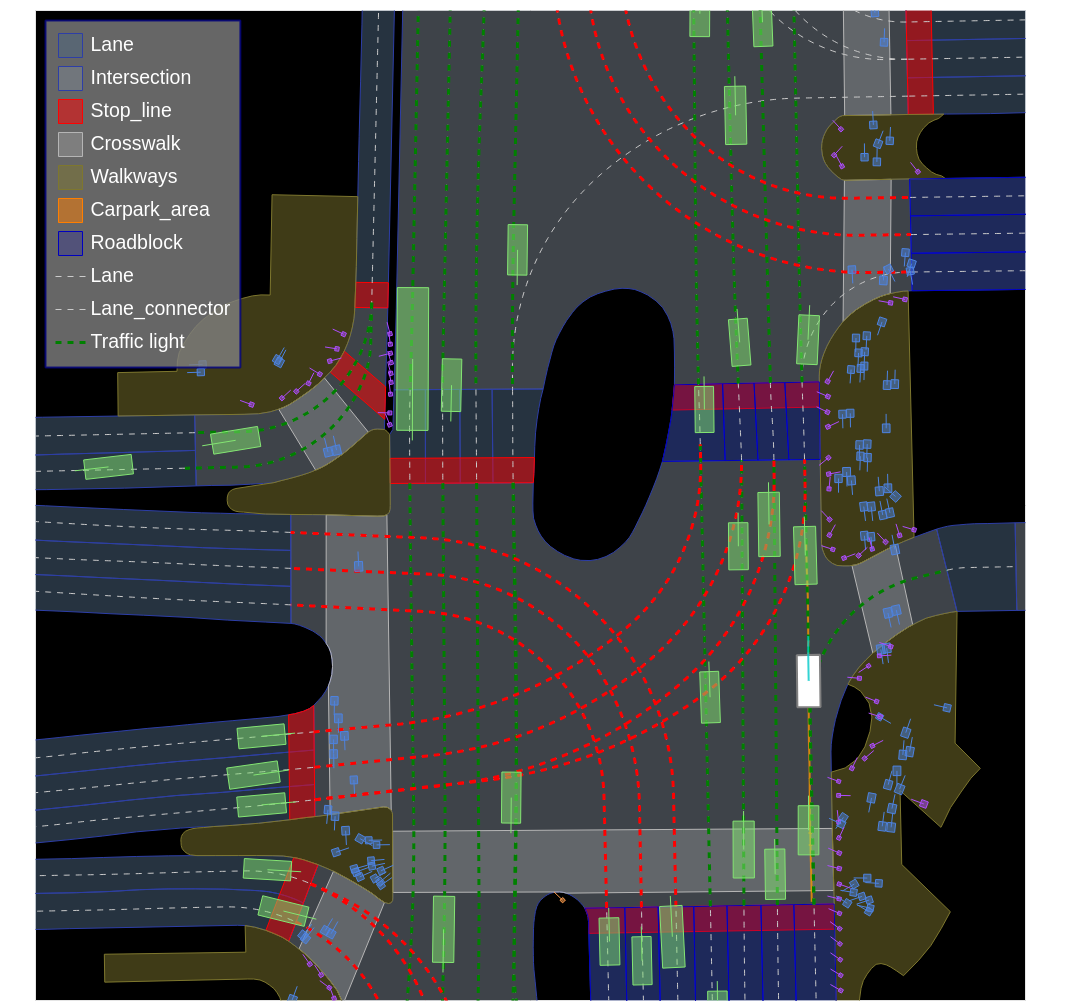}
    \vspace{-2mm}
    \caption{Semantic map of nuPlan with 10 semantic layers in different colors, polygons and lines. It also includes traffic light statuses encoded into the lane connectors.}
    \label{fig:example_semantic_map_layer}
    \vspace{-5mm}
\end{figure}

\textbf{Maps.} 
Similar to nuScenes \cite{nuscenes}, nuPlan provides detailed human-annotated 2D high-definition semantic maps of the driving locations.
We release rasterized and vectorized maps.
While rasterized maps are useful for simplicity and efficient lookup, vectorized maps provide more precise geometric information and metadata.
Examples of semantic map layers are lanes, car parks, crosswalks and stop lines (see Fig.~\ref{fig:example_semantic_map_layer}). 

\subsection{Auto-labeling} 
\label{sec:dataset:autolabeling}
In order to faithfully reconstruct various driving scenarios, we develop an auto-labeling system. It first generates the tracks for all the objects in the scene; then traffic light statuses are inferred from these tracks. Based on the above labels, we can reliably mine different driving scenarios.

\textbf{Offline perception.}
We build an offline perception system to label the objects in the scene~\cite{waymo_offboard} automatically. 
Compared with the online perception systems used in many other datasets~\cite{lyft,argoverse}, the offline version is not constrained by latency and causality. 
Therefore, the fidelity of the generated tracks is drastically higher, enabling us to evaluate planning performance under very limited perception noise.

Inspired by \cite{waymo_offboard}, our offline perception system contains three stages: 3D object detection, offline tracking, and global track refinement. 
The detector in the first stage takes the point clouds from both the top lidar and side lidars as input and detects the bounding boxes through a large neural network~\cite{regnet}.
The offline tracker leverages both past and future detections in an extended time window to generate tracks. In the last step, a novel network is developed to load both tracks and points clouds within the tracks to refine the attributes of all the bounding boxes of vehicle class, such as positions, headings, sizes and velocity.

\textbf{Traffic light status.} To create a realistic simulation of the environment, it is crucial to capture the traffic light statuses.
Existing datasets lack traffic lights~\cite{argoverse,argoverse2} or use online vision-based systems to detect their statuses~\cite{waymo,lyft}.
In contrast, we develop a novel offline system to automatically label the statuses of traffic lights by inferring them from the motion of the actors present in the scene. 
Our labeling system is able to cover all lanes with observed agents. 

We make use of the detections and tracks produced by our offline perception system, as well as map information. To infer a green traffic light status within a given intersection, we determine if there are agents moving within the intersection in the direction controlled by the particular traffic light. To infer a red traffic light status, we check for agents slowing down or being stationary in the lane that approaches the intersection and controlled by the particular traffic light.

\textbf{Scenario mining.} Traditional approaches to evaluate planning performance are dominated by monotonous lane following scenarios. To get a fine-grained understanding of the planning performance, we develop a scenario taxonomy and scenario mining algorithms using low-level attributes like vehicle speed and state transitions. 
The attributes can be inferred from offline perception tracks and traffic light statuses.
In total, we have 73 unique scenario types.

%% file: sections/simulation.tex
\section{Simulation}
nuPlan provides a simulation framework (Fig.~\ref{fig:planning_framework}) that is modular and flexible to work with different datasets and setups.
The simulation is initialized with the real-world \textit{observations} captured in the dataset, namely raw sensor data or object tracks.
Given these environment observations, an \textit{agent} model can be used to predict the future trajectories of all agents.
Observations and agent trajectories are passed to a \textit{planner} that predicts the best route for the ego vehicle given the other agents' routes.
Finally, a \textit{controller} converts the intended route into a feasible trajectory.
The simulation can either playback the actions recorded in the dataset (open-loop) or allow the simulation to deviate from the recording by incorporating the ego's actions (closed-loop).
Below are the simulation components in detail.
\vspace{-4.5pt}
\subsection{Agents}
Of the 6 object classes, vehicle, pedestrian, generic object, traffic cone, barrier, bicycle, construction zone sign,
 3 are moving object classes simulated as agents: vehicles, pedestrians, and cyclists. 
Agents are dynamic objects that can move as the scenario evolves. 

A well-known method is to simply propagate agents according to the logged data. 
We refer to these as non-reactive log-replay agents. 
Log-replay agents are used to simulate scenarios both in open-loop and closed-loop, a near-perfect recreation of the recorded data. 
However, closed-loop simulation quickly diverges if the planner decides to take different actions from what is recorded in the log. Thus, in closed-loop simulation agents can be also simulated with the aim to interact with the ego vehicle and with each other by producing novel simulation states that resemble real agent behaviors. We refer to these as reactive agents. Reactive agents are only relevant in closed-loop simulation and by definition open-loop simulation uses non-reactive agents.

We develop reactive agents following the Intelligent Driver Model~\cite{PhysRevE.62.1805} (IDM) policy.
IDM agents are initialized with the initial pose and velocity of the logged agents.
The agents follow the lane center line of the underlying map.
The longitudinal control is dictated by the IDM policy.
This allows the agents to react to the ego's actions, as well as other reactive agents.
In turn, this reduces false collisions and lets scenarios play out for longer.
Note that we only apply this policy to vehicles, while Vulnerable Road Users (VRUs) are replayed from the log.
We choose not to model VRUs as reactive agents as their behavior is often uncooperative and thus hard to model.

\vspace{-4pt}
\subsection{Controller}
In nuPlan, planners provide a trajectory as a sequence of poses in $SE(2)$, without any kinematic feasibility requirement.
This trajectory is assumed to be sampled at specific times in the future according to the simulation configuration. 
To assert kinematic feasibility and prevent users from cheating, we require the use of a controller.
nuPlan provides the flexibility to use any controller, such as perfect tracking, which simply interpolates poses along the planned trajectory. 

We developed a two-stage controller to propagate the simulation in closed-loop.
This controller consists of two parts, a trajectory tracker and a motion model to forward-integrate the simulation. 
We implement a Linear Quadratic Regulator (LQR)~\cite{liu2021simulation,varma2020trajectory} as the tracker. The found optimal control policy is then fed to the second part of the simulation controller, a kinematic bicycle model which is forward-integrated to propagate the simulation state.
Alternatively, we also support different trackers in the two-stage controller, such as an iterative-LQR tracker.

\subsection{Evaluation}\label{sec:metric}
Different metrics and frameworks have been explored for scoring models in prediction and motion planning benchmarks \cite{waymo,hekmatnejad2019encoding,xiao2021rule, houston2021one,mcallister2022control, maierhofer2022formalization,jiang2022efficient,ilievski2020wisebench}. In this paper, we select a set of metrics and design an aggregation method to compare the performance of planners. 
In open-loop, we only evaluate the closeness of the planner generated trajectory to the human-driven trajectory. 
The open-loop metrics are modified Average Displacement Error (ADE), Final Displacement Error (FDE), Average Heading Error (AHE), Final Heading Error (FHE), and Miss Rate (MR) in which we calculate the metrics over different horizons and report their average score. 
For closed-loop, we use a combination of metrics to evaluate lawfulness and compliance with traffic rules consisting of no at-fault collisions, trajectories inside drivable area, no trajectories in lanes belonging to oncoming traffic, not driving above the speed limit and maintaining enough Time To Collision (TTC) with other road users, metrics to evaluate progress towards the goal and measure the rider comfort.
All metrics' scores are normalized to the range $[0-1]$ using thresholds that are selected based on legal requirements and natural human driving. 
A higher score indicates a better performance.

The final score of a planner is computed by averaging the scores for its generated trajectories across all scenarios. The score of a trajectory in a scenario is given by a hybrid weighted average of all metrics' scores.

The rest of the metrics are weighted according to their importance (See Tab.~\ref{tab:metrics}) and then averaged to compute the scenario score as:
\begin{equation*} 
\label{eq:m}
%\vspace{-2mm}
\mbox{\it scenario score} = \hskip -1em \prod_{i \in \scriptscriptstyle {\text{multiplier metrics}}} \hskip -1em score_i \times \sum_{j \in \scriptscriptstyle {\text{average metrics}}}  \hskip -1em weight_j \times score_j
\vspace{-0mm}
\end{equation*}

We define the score for each challenge (open-loop, closed-loop non-reactive, and closed-loop reactive) as the average scenario score across all scenarios for that challenge.

\begin{table} 
    \vspace{5pt}
    \centering
    \caption{Planner metrics their and weights}
    \label{tab:metrics}
    \vspace{-3mm}
\begin{tabular}{c|ccc}
\hline
Simulation &
  Metric name &
  \begin{tabular}[c]{@{}c@{}}Multiplier\\ weight\end{tabular} &
  \begin{tabular}[c]{@{}c@{}}Average\\ weight\end{tabular} \\ \hline
\multirow{3}{*}{Open-loop}   & MR within bound              & \{0, 1\}       & -  \\
                             & AHE and FHE within bound     & -             & 2 \\
                             & ADE and FDE within bound     & -             & 1 \\ \hline
\multirow{8}{*}{Closed-loop} & No at-fault collisions       & \{0, 0.5,  1\} & - \\
                           & Drivable area compliance     & \{0, 1\}       & -  \\
                           & Making progress              & \{0, 1\}       & -  \\
                           & Driving direction compliance & \{0, 0.5,  1\} & -  \\
                           & TTC within bound             & -             & 5 \\
                           & Progress along route ratio   & -             & 5 \\
                           & Speed limit compliance       & -             & 4 \\
                           & Comfort                      & -             & 2 \\ \hline
\end{tabular}
    \vspace{-6mm}
\end{table}

%% file: sections/experiments.tex
\section{Experiments}
\label{sec:experiments}
Here we present a number of planning baselines and their results when evaluated on the nuPlan benchmark. 
We analyze how the planning performance is impacted by lower quality perception inputs, as well as how it generalizes to other cities.
Finally, we discuss the new state-of-the-art set by the submissions to the first nuPlan challenge.

\subsection{Planning baselines}
We implement several planning methods that are representative of the literature.

\paragraph{Simple Planner}
The Simple planner has little planning capability. The planner plans a straight line at a constant speed. The only logic of this planner is to decelerate if the current velocity exceeds the max velocity.

\paragraph{IDM Planner}
The Intelligent Driver Model (IDM) planner is essentially an Adaptive Cruise Control (ACC) policy \cite{derbel2013modified}. The planner consists of two parts: path planning and longitudinal control.
The path planning component is a breadth-first search algorithm. It finds a center-line path toward the mission goal extracted from the underlying map structure. The longitudinal control follows the IDM policy. The policy describes how fast the planner should go based on the distance between itself and the closest leading agent. 

\paragraph{Raster ML planner}
Similar to the encoder in~\cite{bansal2018chauffeurnet, hu2023imitation}, the raster planner uses ResNet-50~\cite{rw2019timm} as the backbone to encode features from an ego-centric multi-channel raster representing the ego, the agents and the map. The model directly outputs the final ego trajectory. The planner does not perform any post-processing on the predicted ego trajectory.

\paragraph{UrbanDriver ML Planner}
We adopted an open-loop training variant of the UrbanDriver model \cite{scheel2021clt} as a representative machine learning planner baseline. The model processes vectorized agents and map inputs into local feature descriptors that are passed to a global attention mechanism for yielding a predicted ego trajectory. We train the model using imitation learning to match expert trajectories available in the nuPlan dataset. Data augmentation is additionally performed on the agents and expert trajectory provided during training to mitigate data distribution drift encountered during closed-loop simulation. This version was used for the challenge. We also implemented a multi-step prediction baseline variant as discussed in \cite{scheel2021clt} and originally proposed in \cite{Venkatraman_Hebert_Bagnell_2015} to further address the distribution shift, for the experiments in this work but do not open-source this implementation.
\vspace{-1mm}
  \begin{table} 
    \vspace{5pt}
    \centering
    \caption{Main results}
    \vspace{-2mm}
    \label{tab:main_results}
    \begin{tabular}{ c|c c c }
        \bottomrule
        \multirow{2}{4em}{Planner} & \multirow{2}{5em}{Open-loop} & Closed-loop & Closed-loop \\
        &&Non-reactive&Reactive\\
        \hline 
        Simple Planner & 0.22 & 0.32 & 0.37\\
        IDM Planner & 0.30 & \textbf{0.73} & \textbf{0.76}\\
        Raster Planner & 0.52 & 0.47 & 0.46\\
        UrbanDriver & \textbf{0.90} & 0.68 & 0.67\\
        \toprule
    \end{tabular}
    \vspace{-7mm}
\end{table}
\subsection{Main results}
Tab.~\ref{tab:main_results} shows the planning results for the proposed baselines in each of the three challenge setups. Supervised learning-based planners excel in an open-loop setting. This is unsurprising as the task is akin to the traditional motion forecasting challenge. This suggests that an ML planner can choose to make similar decisions to a human driver in open-loop settings.
%the same decisions as a human in isolated instances 
%\holger{I don't agree with that. What we measure is not about being exactly ``the same'' and its not about ``isolated instances''.}. \pat{In open-loop every sample is an isolated instance. There is no causal relationship between the predicted trajectory and the world state. We are using the expert as the GT so we are measuring how close the planner can make the same decision as the person driving the car at the time.} 
However, ML planners still struggle to overcome the distribution shift in closed-loop. A closed-loop scenario can develop into a new situation that was never present in the training dataset.
Even techniques such as data augmentation and closed-loop training fail to overcome this domain gap. This is evident in both the literature \cite{bansal2018chauffeurnet} and our experiments.
Rule-based planners, on the other hand, face no such issues. 
Policies like IDM can produce decent driving behavior. This is confirmed by the metrics as it achieved the highest scores for closed-loop.
It should be noted though that the reactive agents are also modelled with a similar IDM.
The use of similar assumptions on the vehicle behavior may result in giving the IDM planner an unfair advantage over other planners in closed-loop evaluation.
It is evident that sufficiently sophisticated rule-based planners still outperform purely learned planners in closed-loop settings.

\subsection{Perturbation}
As the nuPlan dataset is created with offline perception, to capture the original probability distribution of data collected online, we injected uniform noise on the detections. Noise was added in the dimensions and the pose of the detected agents, with variance extracted by comparing offline and online detections. The scores of planners under nominal and noise-injected simulations in closed-loop reactive mode are presented in Tab.~\ref{tab:detection_perturbation}.
A version of UrbanDriver trained on the perturbed data is called UrbanDriverOnline, which shows a performance deterioration compared to the nominal model on both nominal and injected data. This indicates the value of high-quality offline annotations in the dataset and the learning pipeline.

\begin{table} \vspace{5pt}
    \centering
    \caption{Detection perturbation in closed-loop reactive simulation}
    \vspace{-3mm}
    \label{tab:detection_perturbation}
    \begin{tabular}{ c|c c }
        \bottomrule
        Simulation Agents & UrbanDriver & UrbanDriverOnline \\
        \hline
        Original &  \textbf{0.67} & \textbf{0.60} \\
        Perturbed & 0.64 &  0.59 \\
        \toprule
    \end{tabular}
     \vspace{-2mm}
\end{table}

\subsection{Generalization}
The location generalization experiment is shown in Tab.~\ref{tab:location_generalization}. The experiment aims to test a model's generalization capabilities. The UrbanDriver model was trained purely on data from Las Vegas. The model was tested separately on scenarios from Singapore, Boston, Pittsburgh, and Las Vegas.
The open-loop performance dropped by 53.8\%, while closed-loop non-reactive and closed-loop reactive performance dropped by 35.1\% and 41.5\% respectively. The worst-performing location is Singapore. This can be explained by the left-hand traffic, while the model was trained on right-hand traffic.
One insight is that the correlation between the model's open-loop and closed-loop performance is relatively weak. 
The difference between open and closed-loop scores across Singapore, Boston, and Pittsburgh is only 16.3\%, while for Las Vegas it is more than double at 37.3\%. 
This indicates that a good motion forecasting model does not translate to closed-loop capabilities. 
Thus a major challenge is to overcome the domain gap between open-loop and closed-loop before tackling larger generalization problems.

\begin{table}  \vspace{-2pt}
    \centering
    \caption{UrbanDriver Location Generalization}
    \vspace{-5pt}
    \label{tab:location_generalization}
    \begin{tabular}{ c|c c c }
        \bottomrule
        \multirow{2}{5em}{Locations} & \multirow{2}{5em}{Open-Loop} & Closed-Loop & Closed-Loop \\
        &&Non-reactive&Reactive\\  
        \hline
        Las Vegas & \textbf{0.91} & \textbf{0.57} & \textbf{0.57}\\
        Singapore & 0.28 & 0.23 & 0.18 \\
        Boston & 0.57 & 0.49 & 0.50 \\
        Pittsburgh & 0.41 & 0.39 & 0.32 \\
        \toprule
    \end{tabular}
    \vspace{-6mm}
\end{table}

\subsection{nuPlan challenge}

In the nuPlan motion planning challenge contestants create a planner to traverse a set of diverse and challenging scenarios across all four cities. 
Tab.~\ref{tab:challenge_results} shows the Overall Score, which is the average across Open-loop, Closed-loop Non-reactive, and Closed-loop Reactive challenges of the top four planners. In the open-loop challenge, planners that incorporated supervised learned methods scored relatively well. 
In the closed-loop challenges, planners employed a combination of learned and handcrafted components. 
A common theme was the use of a learned model to first predict the ego's planned trajectory. 
\cite{hu2023imitation} uses a raster-based model that outputs a spatial-temporal heatmap for the ego and an occupancy map for the surrounding agents. 
\cite{huang2023gameformer} are vector-based using transformers as a backbone.
Once the trajectory is obtained, the planner has a further refinement stage used to ensure kinematic feasibility and collision avoidance.
The highest-scoring planner in closed-loop was mostly rule-based \cite{dauner2023parting}.
It generates a handful of trajectories by perturbing the center line laterally at different velocities. 
Trajectories are selected with a heuristic that considers factors such as collision, drivable areas, traffic laws, and comfort. An ML-generated trajectory is fused to correct the long-term planned horizon. This limited the influence of the learned model.
We draw two conclusions from the challenge results.
First, ML-based methods require additional post-processing for closed-loop driving. 
Second, hybrid methods appear to be the most effective approach, combining traditional and data-driven methods.

\begin{table}  \vspace{5pt}
    \centering
    \caption{nuPlan Challenge Leaderboard}
    \label{tab:challenge_results}
    \vspace{-5pt}
    \begin{tabular}{ c | c c }
    \bottomrule
    Team name & Approach & Overall Score\\
    \hline
    CS Tu \cite{dauner2023parting}& Rule-based + ML refinement & \textbf{0.895}\\
    AutoHorizon \cite{hu2023imitation} & ML + optimizer & 0.875\\
    Pegasus & ML + collision checking & 0.848\\
    AID \cite{huang2023gameformer} & ML + hierarchical game theory & 0.829\\
    \toprule
    \end{tabular}
    \vspace{-5mm}
\end{table}

%% file: sections/conclusion.tex
\section{Conclusion}
We presented nuPlan, the first real-world driving benchmark and the largest existing labeled autonomous driving dataset.
The dataset consists of 1282 hours of diverse driving scenarios across 4 cities as well as an unprecedented 128 hours of raw sensor data and is accompanied by an evaluation framework powered by a closed-loop simulator; the dataset and the evaluation framework are publicly available.
We investigated the state of current rule-based and learned-based planners by evaluating multiple approaches on the nuPlan dataset across challenging driving scenarios.
The first public nuPlan challenge demonstrated that rule-based planners outperform purely ML-based ones, but hybrid planners with learned-based components show the most promise in handling difficult scenarios.
In the future, we plan to mine for richer long-tail driving scenarios, design scenario-based metrics, provide ML-based planning and agent baselines and explore end-to-end planner training directly from sensor data.

%% file: sections/acknowledgement.tex
\vspace{-10pt}
\section*{Acknowledgement}
We would like to thank the numerous current and former colleagues at Motional who contributed to this work, especially Juraj Kabzan, Edouard Capellier, Abhinav Rai, Xiaoli Meng, Jiong Yang, Lubing Zhou, Abirami Srinivasan, Bing Jui Ho, Hiok Hian Ong, Mitchell Spryn, Taufik Tirtosudiro, Michael Noronha, Vijay Govindarajan, Jeff Wang, Nelly Lyu, Samson Hang, Shashank Chaudhary, Patrick Weygand, Vern Jensen, Abhimanyu Singh, Qiang Xu and Sammy Omari.
Furthermore, many external advisors gave useful feedback, including Kashyap Chitta and Oscar de Groot.

%% file: sections/appendix.tex
In this supplementary material we provide additional information about the creation of the nuPlan dataset, the simulation framework and evaluation protocol, as well as additional experiments.

\section{Dataset}

\begin{figure}[b!]
  \centering
  \includegraphics[width=\linewidth]{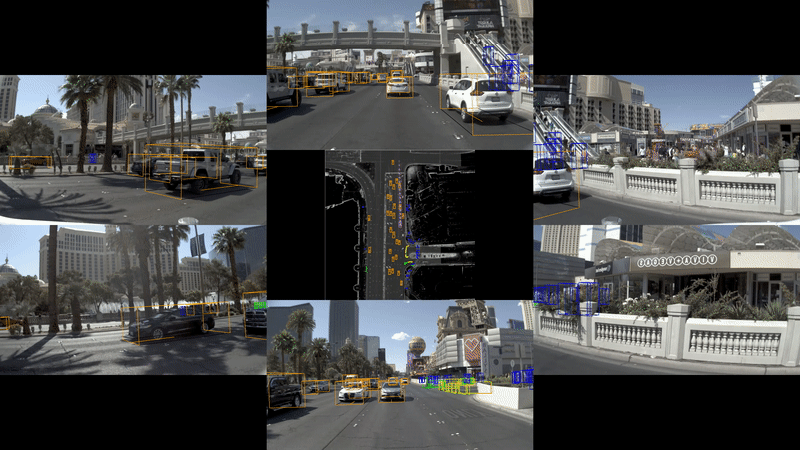}
  \caption{Examples of nuPlan sensor data. We can see the lidar data and map in the center and 6 camera views around it. All views show the highly accurate autolabeling annotations.
  }
  \label{fig:sensor_data_example}
\end{figure}

\subsection{Data collection}
\label{sec:appendix:dataset:data_collection}

Here we describe the sensor setup, synchronization, dataset splits, semantic map layers and object annotation statistics.

\textbf{Sensor setup.}
The nuPlan sensor setup is collected by a fleet of 31 vehicles with identical sensor setup (Tab.~\ref{tab:sensor_setup}).
The vehicles are equipped with 5 lidars, 8 cameras, as well as GNSS \& IMU. 
Thus both lidars and cameras cover the full 360-degree environment and minimize blindspots (Fig.~\ref{fig:sensor_data_example}). The lidars can generate lidar point clouds with up to 360k and 720k points for 20-channel and 40-channel lidars, respectively. 

\begin{table}
    \centering
    \caption{Sensor setup in nuPlan}
    \vspace{-3mm}
    \resizebox{\linewidth}{!}{
    \begin{tabular}{L{0.12\linewidth} | L{0.10\linewidth} | L{0.08\linewidth} | L{0.40\linewidth}}
    \bottomrule
    Sensor & Make & Freq. & Details \\
    \hline
    1x Top Lidar & Pandar 40P & 20Hz & Spinning, 40 channels, 360$^{\circ}$ horizontal FOV, $\leq$ 200m range, $\leq$ $\pm$ 2cm, 40$^{\circ}$ ([-25$^{\circ}$, 15$^{\circ}$]) vertical FOV, $\leq$ 720k points per second \\
    \hline 
    2x A-Pillar Lidars & Pandar 40P & 20Hz & Spinning,  40 channels, 360$^{\circ}$ horizontal FOV, $\leq$ 200m range, $\leq$ $\pm$ 2cm, 40$^{\circ}$ ([-25$^{\circ}$, 15$^{\circ}$]) vertical FOV, $\leq$ 720k points per second\\
    \hline
    2x Bumper Lidars & Pandar 20PA front, 20PB rear & 20Hz & Spinning, 20 channels, 360$^{\circ}$ horizontal FOV, $\leq$ 200m range, $\leq$ $\pm$ 2cm, 33$^{\circ}$ ([-25$^{\circ}$, 8$^{\circ}$]) front vertical FOV, 22$^{\circ}$ ([-19$^{\circ}$, 3$^{\circ}$]) rear vertical FOV, $\leq$ 360k points per second\\
    \hline
    8x Cameras & D3 Engineering D3RCM & 10Hz & RGB, $1/2.7\textrm{''}$ CMOS sensor, 2000x1200 resolution, split-pixel image sensor\\
    \hline
    1x GNSS & Trimble BX992 & 20Hz & Position latency $<$ 20ms, 0.10$^{\circ}$ roll/pitch \\
    \hline
    1x IMU & Honey- well HG1120 & 100Hz & MEMS gyroscopes, accelerometers and magnetometers\\
    \toprule
    \end{tabular}
    }
    \label{tab:sensor_setup}
\end{table}

\textbf{Synchronization.} 
To ensure high-quality data alignment between multiple sensors, lidar sensors are synchronized with the system time on the car in multiple lidar periods. The exposure of each camera is triggered in the same way plus a camera-specific offset, where the offset is an optimal value to ensure images are acquired right when the lidars are sweeping through each camera's FOV. 
Given there are multiple lidar sweeps in a spin, we merge a full sweep from each lidar sensor into a merged sweep since they complete a revolution at nearly the same instance. After they are merged, the merged point clouds are transformed to the same frame coordinate and timestamp. We also perform ego motion compensation to take ego position into account when the points are acquired. 

\textbf{Dataset splits.}
The dataset splits (train, val, and test) are geographically overlapping, which means they can cover the same region in a city.
However, to minimize temporal data leakage, data from the same day \emph{and} city is not shared across splits.
In addition, the dataset is stratified across days, cities, and driving scenarios to ensure all splits have similar balances across these dimensions.

\textbf{Map layers.}
We define the semantic map layers in nuPlan. The map layers are designed with planning in mind. A graph can be constructed 
from the network of lane and lane connectors. Additionally, the map is annotated with important road features (stop lines, crosswalks, car parks) that impact the decision-making of the AV. See Tab.~\ref{tab:map_layers} for more details.

\begin{table}
    \centering
    \caption{Semantic map layers in nuPlan}
    \label{tab:map_layers}
    \resizebox{\linewidth}{!}{
    \begin{tabular}{L{1.8cm} | L{6cm}}
      \bottomrule
      Layer name & Layer description\\
      \hline
      Baseline paths & The center line along lanes and lane connectors.\\
      \hline
      Carpark areas & Polygons representing car parks. \\
      \hline
      Crosswalks & Polygons representing crosswalks. \\
      \hline
      Generic drivable areas & Polygons representing areas where the AV is allowed to drive.\\
      \hline
      Intersections & An area connecting multiple road segments together. \\
      \hline
      Lane connectors & Road segments connecting two lanes. Typically, they exist within an intersection.\\
      \hline
      Lane group connectors & A group of adjacent lane connectors that travel in the same direction\\
      \hline
      Lane groups polygons & A group of adjacent lanes that travel in the same direction.\\
      \hline
      Lanes polygons & Polygons representing a lane.\\
      \hline
      Road segments & Groupings of lane groups that travel in opposite directions from one another.\\
      \hline
      Stop polygons & Polygon in which the AV may be required to stop.\\
      \hline
      Traffic lights & 3D locations of a set of traffic lights.\\
      \hline
      Walkways & Polygons representing walkways.\\
     \hline
    \toprule
    \end{tabular}
    }
\end{table}

\begin{figure}
    \centering
    \includegraphics[width=0.40\textwidth]{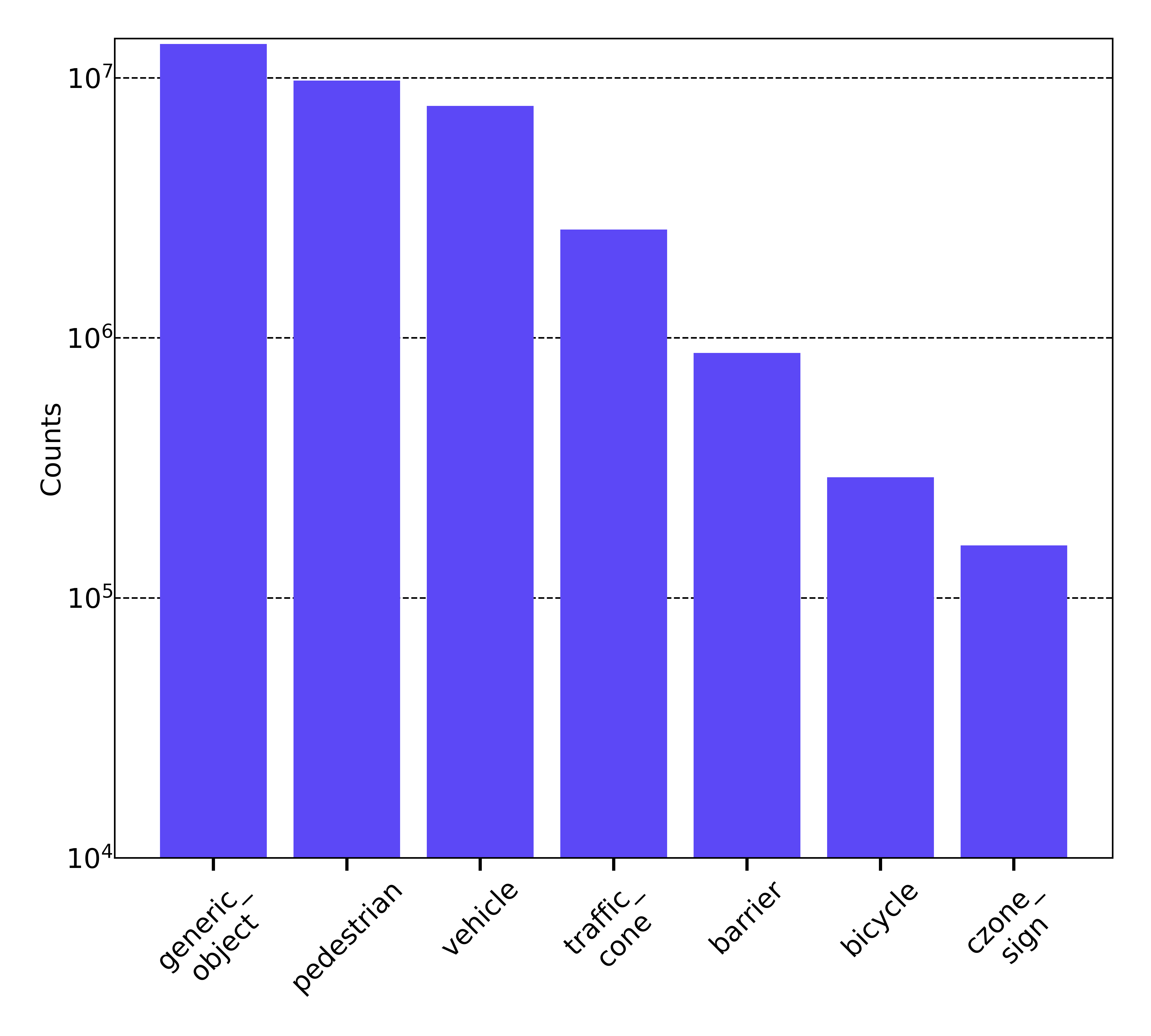}
    \caption{Number of tracks per category.}
    \label{fig:nuplan_box_counts_SM}
\end{figure}

\textbf{Annotation statistics.}
Here we present more statistics for the annotations in nuPlan.
Fig.~\ref{fig:nuplan_box_counts_SM} shows the number of tracks for each of the 6 object classes in nuPlan. 
While there are about 2 orders of magnitude between the most (generic object) and least (construction zone sign) common class, the dataset is nevertheless less long-tailed than other datasets with more fine-grained classes~\cite{nuscenes}.
Furthermore, even the rarest class still has more than $10^5$ tracks, which shows the advantage of such a large dataset.

\begin{figure*}
  \centering
  \includegraphics[width=\textwidth]{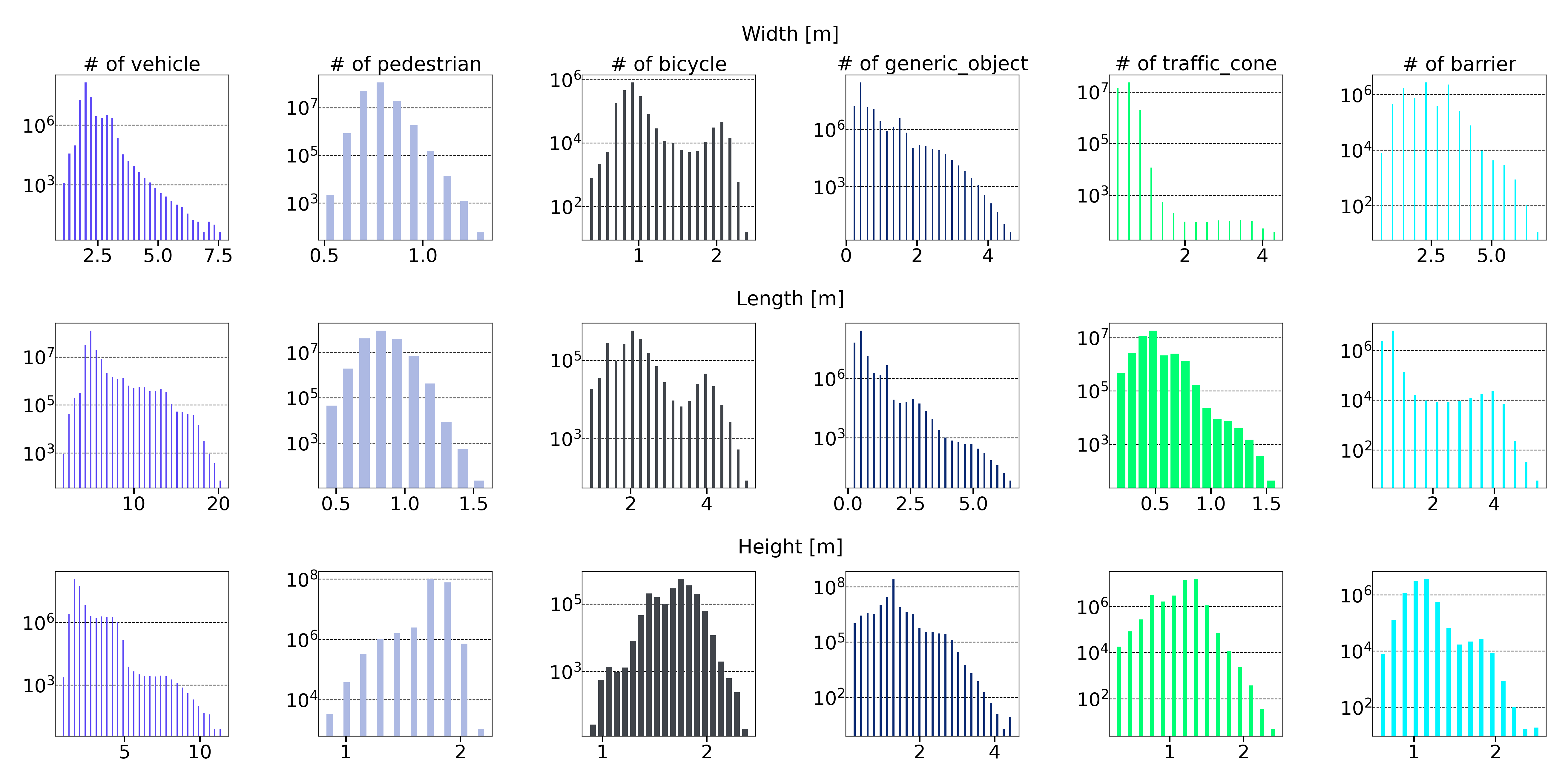}
  \caption{Size distributions of boxes.}
  \label{fig:nuplan_box_size_distributions_SM}
\end{figure*}

We also present size statistics for different classes in Fig.~\ref{fig:nuplan_box_size_distributions_SM}. 
These show that our boxes are statistically stable in their sizes except for the vehicle class that includes construction vehicles with irregular shapes.

\begin{figure}
    \centering
    \includegraphics[width=\linewidth]{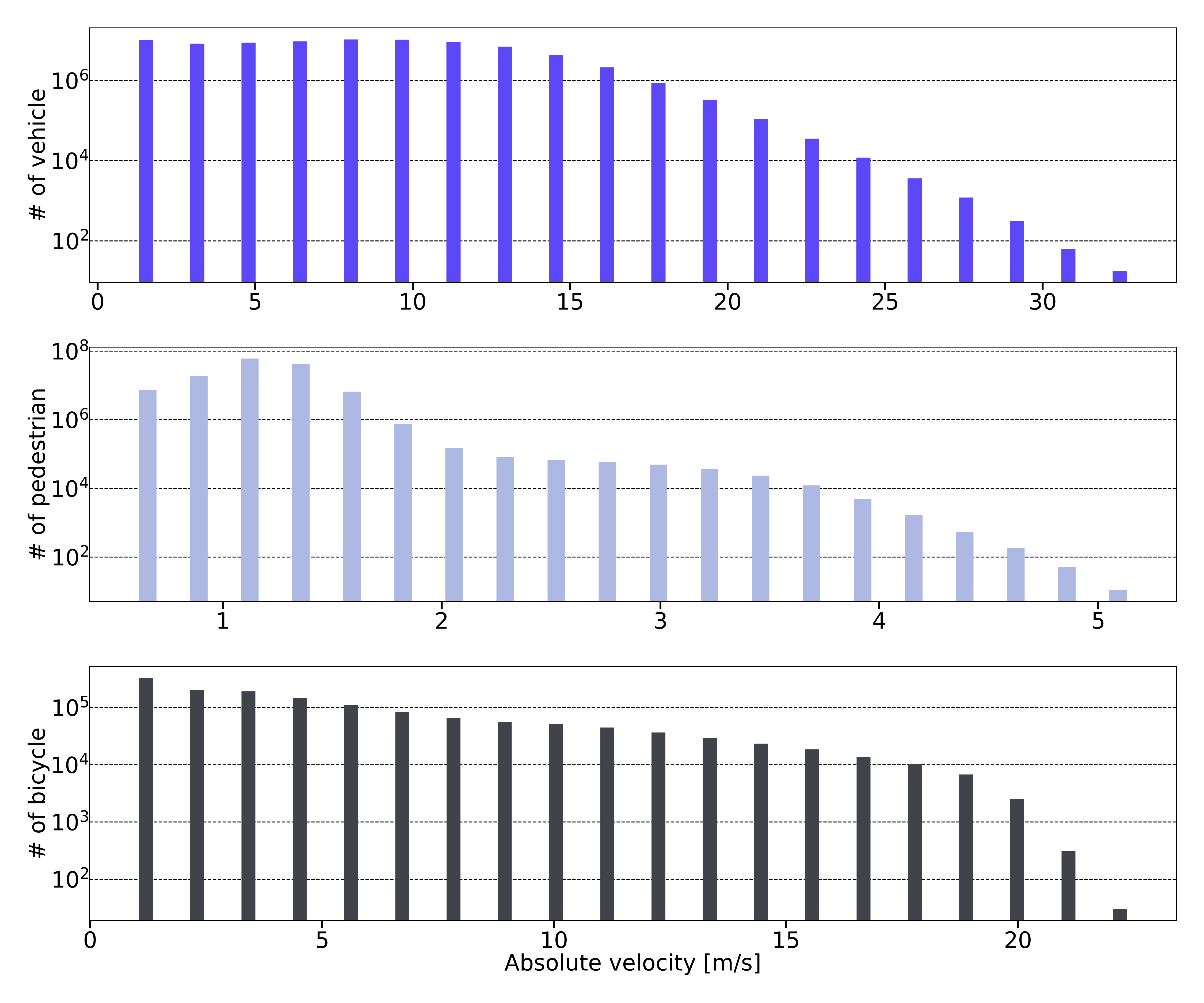}
    \caption{Histogram of the absolute box velocities per class. 
    Only boxes with speed $> 0.5m/s$ are shown. 
    Note the logarithmic scaling on the y axis.}
    \label{fig:box_absolute_velocity_SM}
\end{figure}

Fig.~\ref{fig:box_absolute_velocity_SM} shows the absolute velocities for three classes (vehicle, pedestrian, and bicycle), since other classes are static most of the time. 
We observe that most of the objects are slow-moving, which represents primarily urban scenarios, while a small number of pedestrians and bicycles are moving at unrealistically high speeds, possibly due to noisy annotations.

\begin{figure}
    \centering
    \includegraphics[width=\linewidth]{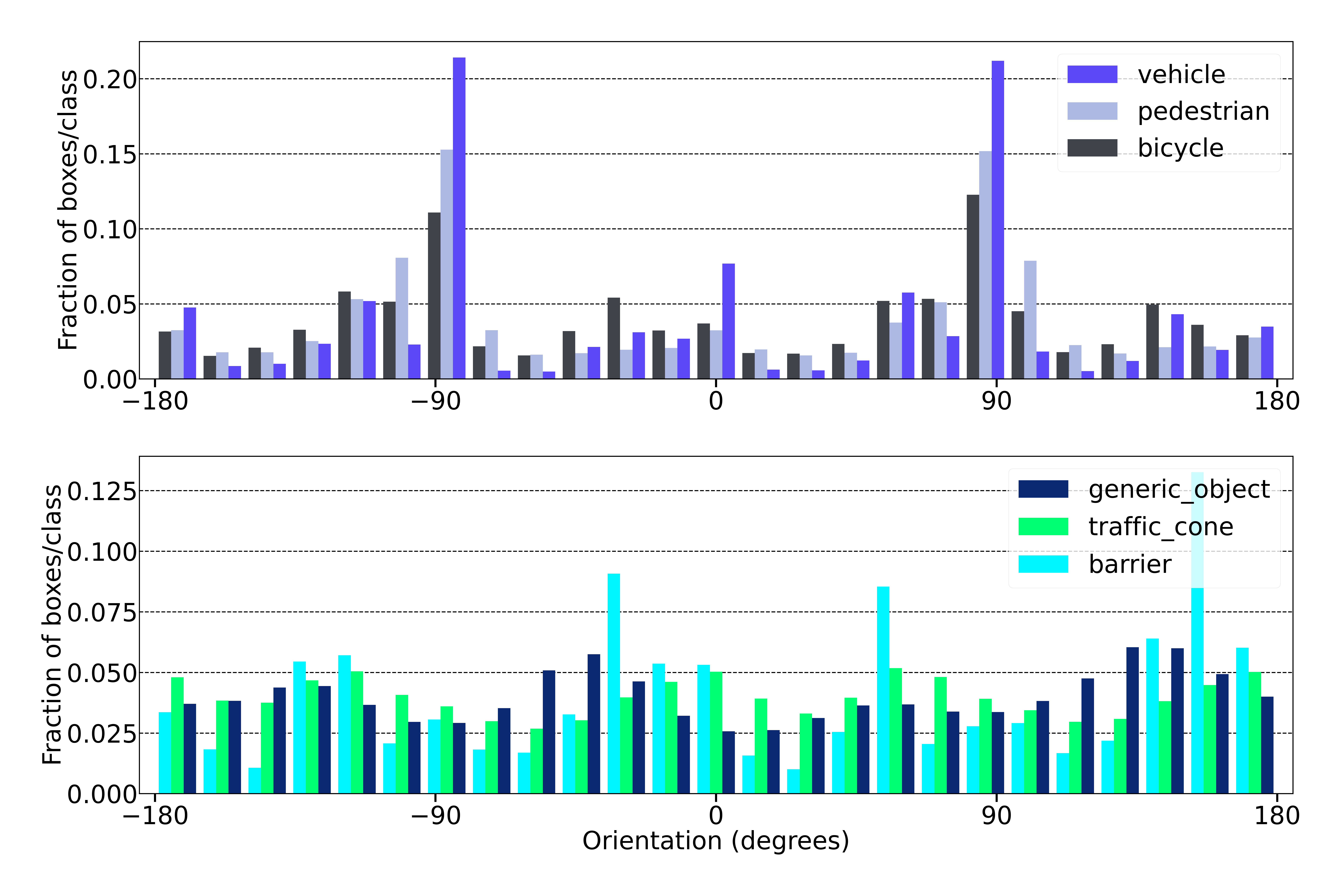}
    \caption{Orientation of boxes.}
    \label{fig:box_orientation_SM}
\end{figure}

Fig.~\ref{fig:box_orientation_SM} shows the object (box) orientations relative to the ego vehicle orientation.
Due to the grid-like road layout, most vehicles have orientations that are multiples of $\pm~90^{\circ}$.

\begin{figure}
    \centering
    \includegraphics[width=\linewidth]{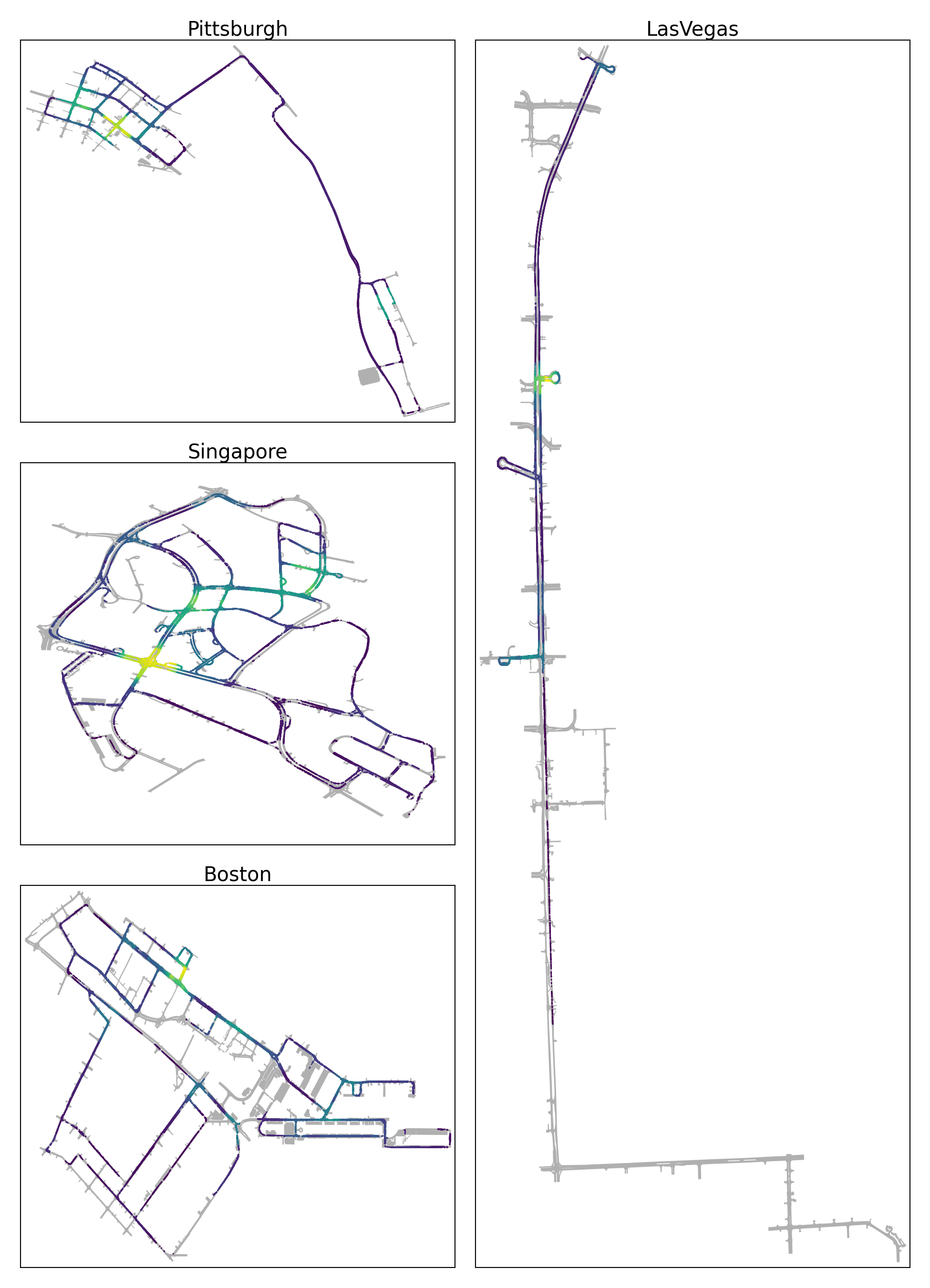}
    \caption{Spatial data coverage for four cities in nuPlan. 
    Colors indicate the number of scenarios with subsampled ego vehicle poses within a 100m radius. 
    There is more ego vehicles in a location as the colors become more yellow.}
    \label{fig:ego_spatial_coverage_SM}
\end{figure}

Fig.~\ref{fig:ego_spatial_coverage_SM} shows the spatial coverage of our ego vehicles across all maps. 
In Las Vegas, most routes start and end in PUDOs, which leads to these areas being visited more often.
In other cities, the distribution is more uniform, with key intersections being visited the most.

\subsection{Autolabeling}
\label{sec:appendix:dataset:autolabeling}
As described in the main paper, We developed an offboard perception system to generate the bounding boxes and tracks for the objects in the scene.
It consists of three stages: object detection, offline tracking, and global track refinement.
The system is deployed on the raw sensor data from nuPlan dataset. 

\textbf{Object detection.}
For the first stage, we extend the state-of-the-art MVF++~\cite{waymo_mvf,waymo_offboard} as our lidar-based 3D object detector.
We select $n$ past frames ($n=7$) and compensate the points based on ego-motion, which significantly densifies the point clouds in the scene.
We find that there is no improvement in incorporating future frames as well.
To further boost the performance, we make several changes to the architecture.
First, different from other works~\cite{waymo_offboard,wang2020pillar}, we propose to include the points from four side lidars in addition to the top lidar, so the objects closer to the ego vehicle will not be missed. 
To use these points properly, we adopt the cylindrical view~\cite{wang2020pillar} in MVF++, instead of the perspective view, to mitigate the collision of points from different lidars.
Second, we enlarge the model capacity by using a RegNet~\cite{regnet} backbone.
The RegNet backbone processes voxelized outputs from MVF++ and generates expressive feature maps for final detection heads.
Third, a CenterPoint~\cite{centerpoint} detection head is applied to predict the bounding boxes.
Since the resulting object boxes are often flipped, we perform majority voting using past and future detections and adjust the heading accordingly.

\textbf{Offline tracking.}
Given the instantaneous bounding boxes generated from the first stage, a Kalman filter based multi-object tracker~\cite{weng2020tracking} is applied to associate the boxes across timestamps.
Since the tracker is running in offline mode, we leverage more information from both past and future frames to manage tracks:
First, after $Bir_{min}$ frames of detections are used to confirm a track, the track is initiated from the first frame of $Bir_{min}$ frames, instead of the last one~\cite{weng2020tracking}. 
As a result, the number of False Negatives is reduced.
Second, compared with online tracking~\cite{weng2020tracking}, we increase $Age_{max}$ to maintain a longer memory before coasting a track.
While a higher $Age_{max}$ slows down the method, the number of ID switches is reduced significantly.

\textbf{Global track refinement.}
Even though the detection network takes $n$ frames of lidar point clouds as input, it still perceives the scene in a short time window.
The resulting points are limited to a certain perspective, posing challenges to estimating the size and heading accurately.
Moreover, although the bounding boxes from the detection network have been smoothed to some extent by a Kalman Filter based tracker, the smoothing does not happen at a global scale.
To mitigate this effect, we introduce a novel neural network. 
Inspired by recent works~\cite{waymo_offboard,auto4d}, our network loads tracks along with the aggregated point clouds within the tracks and refines the bounding boxes in the tracks in terms of position, heading, size and velocity.
Notably, instead of using two networks for static and dynamic objects~\cite{waymo_offboard}, or for different purposes (size, position or heading)~\cite{auto4d}, we design a single network, which achieves the above goals in one shot.
In practice, the proposed method reduces the deployment overhead compared to other works~\cite{waymo_offboard,auto4d}, because the order of the two networks is not clear and the classification of dynamic/static is arbitrary for some slow-moving objects. 
Similar to these works, we also only apply this network to the vehicle class.

\textit{Implementation Details.}
For simplicity, we consider objects on a 2D bird's eye view plane only and denote $(x, y)$ as the object's position, $(w, l, h)$ as size, $(v_x, v_y)$ as velocity and $\theta$ as heading.
An object at timestamp $t$ in a track is denoted as $\mathfrak{O}_t$. 
It has a bounding box $\mathfrak{B}_t = (x_t, y_t, z_t, w_t, h_t, l_t, \theta_t)$ that tightly contains the points $\mathcal{P}_t$.
We select the boxes from the past and future $n_b$ frames. 
These $(2n_b+1)$ bounding boxes are concatenated as the input for the trajectory encoding branch. 
Meanwhile, the point clouds are cropped in both past and future $n_p$ frames and then transformed to the current frame $t$. 
The aggregated point cloud is processed by dynamic voxelization~\cite{waymo_mvf} and taken as input by the point encoding branch. 
Convolutional layers and global average pooling are applied for each branch to generate expressive features, which are concatenated for the final MLP. In training, we use a smooth L1 loss to regress the residuals between the ground truth and input $\mathfrak{B}_t$.
In deployment, after getting the refined bounding boxes for the whole track, we choose the median value of the size for all the bounding boxes.
We update the position and velocity of each box according to the outputs of the network.

\subsection{Evaluation of the autolabeling system}

\label{sec:appendix:experiments:autolabeling}
% Dataset
\textbf{Dataset.}
To evaluate the proposed autolabeling system we use an internal dataset that follows a similar data distribution to nuPlan and contains 3098 human-labeled scenes collected from Singapore, Boston, Pittsburgh and Las Vegas.
Among them, 2958 scenes are used for training, 49 for validation and 91 for testing.

% Evaluation
\textbf{Metrics.}
The performance of the detection network and the global track refinement are evaluated using the maximum F1 scores for all the classes.
The performance of the offline tracker is evaluated using AMOTA, ID switches, recall and F1 score.
Vehicles are divided into two categories depending on whether their lengths are larger than 7m.
We use Birds Eye View 2D IOU as the matching criterion.
The IOU threshold is $0.7$ for vehicles, $0.5$ for cyclists and pedestrians, $0.3$ for barriers and generic objects, $0.15$ for traffic cones. 

\textbf{3d object detection.}
% Baseline
The detection network proposed in this paper is compared with a modified CenterPoint~\cite{centerpoint} detector, termed \emph{mCenterPoint}.
Our goal is to compare our proposed network with a detection network often deployed onboard for autonomous vehicles.
We choose mCenterPoint as a baseline because it is lightweight and meets the real-time requirement. It differs from our offline detection network in three aspects. First, it aggregates past 3 sweeps of point clouds as input to the network. Second, mCenterPoint uses the vanilla PointPillars~\cite{pointpillars} for voxelization. Third, a more compact backbone is used to extract features for the CenterPoint detection heads. In contrast, the offline model is applied on 7 sweeps of point clouds with a multi-view encoder~\cite{waymo_offboard} and a heavy backbone~\cite{regnet}.
% Results
The max F1 scores for each class are shown in Tab.~\ref{tab:detection_exp}.
From the table, we can see the proposed detection network can outperform mCenterPoint by a large margin. It clearly shows that using more sweeps of point clouds as well as a more complex network architecture improves the performance drastically.

\begin{table}
    \centering
    \caption{Detection performance (max F1) of the proposed autolabeling detector, compared to a modified version of the commonly used CenterPoint detector
    }
    \vspace{-3mm}
    \begin{tabular}{ccc}
    \bottomrule
    Class          &   mCenterPoint    &   Ours            \\
    \hline
    Short vehicle   &   72.0            &   \textbf{86.2}   \\
    Long vehicle    &   63.9            &   \textbf{74.8}   \\
    Cyclist         &   64.3            &   \textbf{73.0}   \\
    Pedestrian      &   59.5            &   \textbf{75.1}   \\
    Traffic cone    &   68.7            &   \textbf{78.3}   \\
    Barrier         &   60.0            &   \textbf{68.3}   \\
    Generic object  &   60.2            &   \textbf{68.8}   \\
    \hline
    Average         &   64.1               &    \textbf{74.9}\\
    \toprule
    \end{tabular}
    \label{tab:detection_exp}
\end{table}

\textbf{Offline tracker.}
% Baseline
We also implemented a modified version of AB3DMOT~\cite{weng2020tracking} as a baseline in the tracking experiments. 
Similar to mCenterPoint, \emph{mAB3DMOT} is configured for real-time deployment onboard for autonomous vehicles. In particular, $Bir_{min}$ and $Age_{max}$ are set to be $3$ and $0.4s$ for \emph{mAB3DMOT}, respectively. In comparison, the offline tracker uses $Age_{max}=2s$ and back-traces the first 2 frames with $Bir_{min}=3$.
% Results
From Tab.~\ref{tab:tracker_exp} we see a significant improvement on all tracking metrics.
In particular, the number of ID switches has been reduced by $78\%$, which is due to the significantly enlarged $Age_{max}$. The fewer number of ID switches indicates that the tracks in nuPlan dataset are less fragmented and thus reflect the agents' movement in the real world.  

\begin{table}
    \centering
    \caption{Comparison of tracking metrics between the proposed offline tracker and the modified AB3DMOT}
    \vspace{-3mm}
    \begin{tabular}{@{}ccccc@{}}
    \bottomrule
    Method   & AMOTA          & ID switch     & Recall         & F1 score        \\ 
    \hline
    mAB3DMOT & 0.611          & 9779          & 0.696          & 0.828          \\
    Ours     & \textbf{0.684} & \textbf{2129} & \textbf{0.727} & \textbf{0.868} \\ \bottomrule
    \end{tabular}
    \label{tab:tracker_exp}
\end{table}

\textbf{Global track refinement.}
% Baseline
To evaluate global track refinement, we compare it against the outputs of the offline tracker.
Because the offline tracker will interpolate and suppress some bounding boxes from the detection network, the F1 scores reported here are different from Tab.~\ref{tab:detection_exp}.
% Results
The results are presented in Tab.~\ref{tab:trn_exp}. It is shown that the vehicle F1 score is consistently improved by global track refinement.
In particular, when a stricter matching criterion is applied (BEV IOU = $0.9$), we see a relative boost of $56.8\%$, which demonstrates the efficacy and necessity of global track refinement in producing high-quality bounding boxes.
\begin{table}
    \centering
    \caption{Comparison of F1 scores of vehicle bounding boxes between the proposed global track refinement and the offline tracker.}
    \vspace{-3mm}
    \begin{tabular}{@{}ccc@{}}
    \bottomrule
    Method                  & IOU $0.7$ & IOU $0.9$ \\
    \hline
    Offline-tracker         & 83.4          & 21.3          \\
    Global track refinement & \textbf{87.5} & \textbf{33.4} \\ \bottomrule
    \end{tabular}
    \label{tab:trn_exp}
\end{table}

\subsection{Traffic lights}
\label{sec:appendix:dataset:traffic_lights}
To automatically label traffic light statuses within nuPlan, we make use of the tracks produced by our autolabeling system, as well as the human-annotated map information.
The map indicates the intersections with traffic lights.
Within each traffic light intersection, the individual traffic lights control the flow of vehicles from a lane on one side of the intersection to another lane.
Each such pair of lanes is connected via a \emph{lane connector} in our map.
Hence, we encode the status of the traffic lights in the corresponding lane connectors.

\textbf{Green and amber statuses.}
We consider both green and amber statues of traffic lights to be the same, i.e. green, since vehicles are allowed to move under both light statuses. 
To infer the presence of a green traffic light on a particular lane, we determine if there are agents moving along the lane connector corresponding to the traffic light. 
We do this by comparing the directed Hausdorff distance between the trajectories of all the agents within the traffic light intersection, and the lane connector.
Note that as the traffic light labeling system is offline, we are able to make use of both the past and future trajectories of each agent.
If the directed Hausdorff distance is small, we surmise that there are agents moving along the lane connector, and thus the traffic light controlling that lane connector is likely green.

\textbf{Red statuses.}
To infer the presence of a red traffic light on a particular lane, we check for the minimum speed of the agents that are on that lane.
Note that we only take into account agents which are of a certain distance from the traffic light intersection.
If the minimum speed of the agents on that lane is low, we surmise that the agents are stopped.
We also consider the deceleration of the agents on the various lanes.
When a traffic light is red, it is common for drivers to begin decelerating as they approach the intersection.
Thus, we set a threshold for the deceleration magnitude above which we consider the agent in a lane to be decelerating.
If we find that either the agents on a given lane are stopped or decelerating, then we infer that the traffic light controlling that lane connector is likely red.
For lanes and lane connectors for which there are no observable agents, we set the status of the corresponding traffic light to `unknown`.

\textbf{Post-processing.}
After inferring the per-frame green and red statuses, we perform post-processing to refine the inferences. 
First, we perform grouping.
Our map stores information on lane connectors that go in a "parallel" direction, and therefore share the same traffic light statuses.
We use this information to set all lane connectors in the parallel direction to have the same status. This helps to reduce false negatives, especially in situations when there are no observed agents on some of the lanes going in the same direction.

Second, we perform back-filling.
When a traffic light changes from red to green, drivers often have a certain reaction time before moving off.
Similarly, when a traffic light changes from yellow to red, there may be drivers still crossing the intersection.
This might lead to the system inferring a green traffic light for the particular frame, even though the lights have changed in reality.
Hence there is a slight lag in the transitions identified via motion inference with respect to the actual transition.
To account for this, for all green statuses, we go back a specified time horizon into the past and override the statuses of each lane connector by setting them to green.
We do the same for the red statuses and override the past statuses within the specified time horizon with red.

\textbf{Evaluation.}
To perform a more quantitative evaluation of our traffic light labeling system, we select scenes where the ego vehicle is moving through traffic light intersections and manually label the traffic light statuses for a subset of the data.
We select scenes from various cities and various traffic light behaviors (e.g. constant, transition). We manually label about 1000 frames.
We seek to compare our system against a more conventional traffic light detection system.
Such a system is usually deep-learning based and vision-only.
Consequently, for the baseline, we use YOLOv3 \cite{redmon2018yolov3}, as implemented by \cite{sovitTrafficLight} and trained on the LISA Traffic Light Dataset \cite{mortenTrafficLightDataset}, which contains 113,888 annotated traffic lights. 
We do not fine-tune YOLOv3 on any nuPlan data.
By labeling the statuses of the traffic lights via motion inferences, we are able to recover $5.2$x more traffic light statuses than YOLOv3.
We find that YOLOv3~\cite{redmon2018yolov3} misses several traffic lights at long distances.
It also performs poorly at very close distances to the traffic light due to the perspective warping of the cameras.
Using YOLOv3 \cite{redmon2018yolov3}, the traffic light classification accuracy among visible traffic lights is $66.4\%$. 
In contrast, our system performs slightly better with an accuracy of $68.7\%$. 
We find that the transitions of the traffic light statuses are difficult to infer for our system, due to the lag in the agents moving off, slowing down or stopping whenever the traffic lights change. 
A data-driven alternative could provide better results here.

\subsection{Scenario mining}
\label{sec:appendix:dataset:scenario_mining}
\begin{figure}
  \centering
  \includegraphics[width=\linewidth]{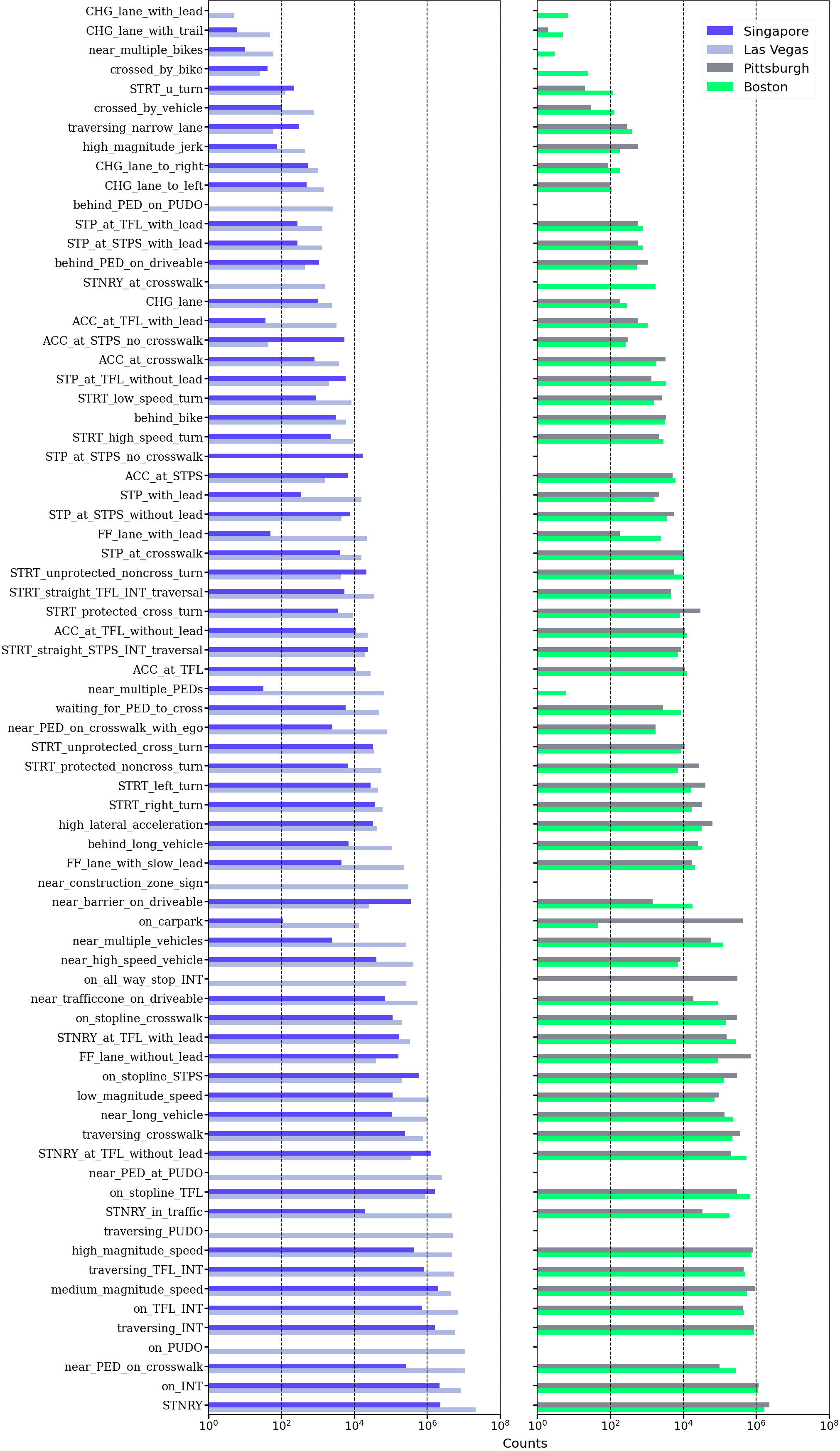}
  \caption{Distribution of scenario types in each city.}
  \label{fig:nuplan_scenario_type_distributions}
\end{figure}

We propose the following approach to mine scenarios from the nuPlan dataset.
First, we compute a large number of atomic primitives.
These primitives model an attribute (vehicle speed) or state transition (a vehicle being in two lanes simultaneously) and can be extracted from the entire dataset in a single pass.
Second, we combine several primitives into an SQL query that can be run efficiently on a database.
Third, we post-process the query results by including a sequence of $10$ seconds before and after the returned time step.
Finally, we manually QA $100$ examples for each scenario. 
If the false positive rate is below $90\%$, we either refine the query by adding more attributes or tuning hyperparameters or discard the scenario.
This approach results in high-precision scenario labels, while recall is less relevant for us.
We have a total of $73$ unique scenario types in the dataset, and their distributions per city are shown in Fig.~\ref{fig:nuplan_scenario_type_distributions}.
The details and parameters of the 14 scenarios that were used to grade the nuPlan planning challenge can be found in Tab.~\ref{tab:scenario_descriptions}.
\begin{table}
    \centering
    \caption{nuPlan planning challenge scenario details}
    \label{tab:scenario_descriptions}
    \resizebox{\linewidth}{!}{
    \begin{tabular}{L{1.8cm} | L{6cm}}
      \bottomrule
      Scenario type & Scenario description\\
      \hline
      starting straight traffic light intersection traversal & Ego at the start of a traversal going straight across an intersection area controlled by traffic lights while not stopped.\\
      \hline
      high lateral acceleration & Ego high ego acceleration $(1.5 < acceleration < 3 m/s^2)$ across the lateral axis with high yaw rate while not turning. \\
      \hline
      changing lane & Ego at the start of a lane change towards an adjacent lane. \\
      \hline
      high magnitude speed & Ego high velocity magnitude with low acceleration $(velocity > 9 m/s)$. \\
      \hline
      low magnitude speed & Ego low ego velocity magnitude $(0.3 < velocity < 1.2 m/s)$ with low acceleration while not stopped. \\
      \hline
      starting left turn & Ego at the start of a traversal turning left across an intersection area while not stopped. \\
      \hline
      starting right turn & Ego at the start of a traversal turning right across an intersection area while not stopped. \\
      \hline
      stopping with lead & Ego starting to decelerate $(acceleration\ magnitude < -0.6 m/s^2, velocity\ magnitude < 0.3 m/s)$ with a leading vehicle ahead $(distance < 6 m)$ at any area. \\
      \hline
      following lane with lead & Ego following $(velocity > 3.5 m/s)$ its current lane with a moving leading vehicle ahead on the same lane $(velocity > 3.5 m/s, longitudinal\ distance < 7.5 m)$. \\
      \hline
      near multiple vehicles & Ego nearby $(distance < 8 m)$ of multiple $(>6)$ moving vehicles while ego is moving $(velocity > 6 m/s)$. \\
      \hline
      traversing pickup dropoff & Ego during the traversal of a pickup/drop-off area while not stopped. \\
      \hline
      behind long vehicle & Ego behind $(3 m < longitudinal\ distance < 10 m)$ a long $(length > 8 m)$ vehicle in the same lane as ego $(lateral\ distance < 0.5 m)$. \\
      \hline
      waiting for pedestrian to cross & Ego waiting for a nearby $(distance < 8 m, time\ to\ intersection < 1.5 m)$ pedestrian to cross a crosswalk area while ego is not stopped and the pedestrian is not at a pickup/drop-off area. \\
      \hline
      stationary in traffic & Ego is stationary with multiple $(>6)$ vehicles nearby $(distance < 8 m)$. \\
     \hline
    \toprule
    \end{tabular}
    }
\end{table}

\section{Simulation}

\subsection{Agents}
\begin{figure}
    \centering
    \includegraphics[width=\linewidth]{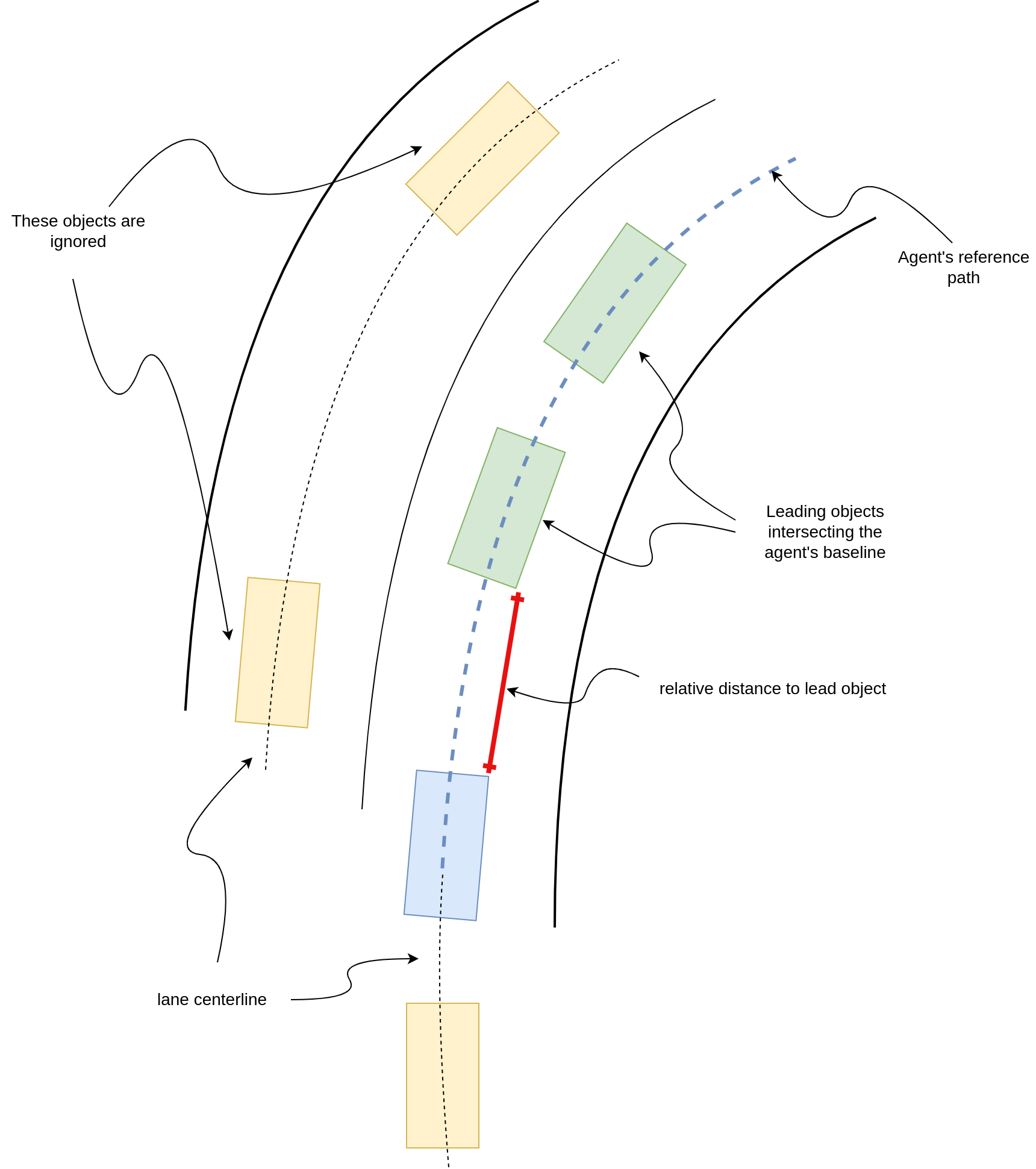}
    \caption{Filtering for all other objects (green) that intersect with the agent's (blue) baseline. The baseline is expanded into a polygon along the path and as wide as the agent's width. All other objects (yellow) are ignored. Checking for the nearest object out of the filtered ones.}
    \label{fig:idm_lead_agent_search}
\end{figure}

The differential equation behind the IDM policy operates based on a focus agent - the vehicle that the IDM policy is controlling - and a lead object. The policy is parameterized by the distance to a lead object. However, in closed-loop simulation identifying the correct lead object is not trivial.
Fig. \ref{fig:idm_lead_agent_search} shows how the closest leading object is identified. Searching for the closest object often returns another agent in the adjacent lane. Hence, the search space is reduced to only the other objects in the scene that are or can potentially intersect with the agent's planned path. The euclidean distance between the two closest points of the focus agent and the lead object is computed.

\begin{figure}
    \centering
    \includegraphics[width=\linewidth]{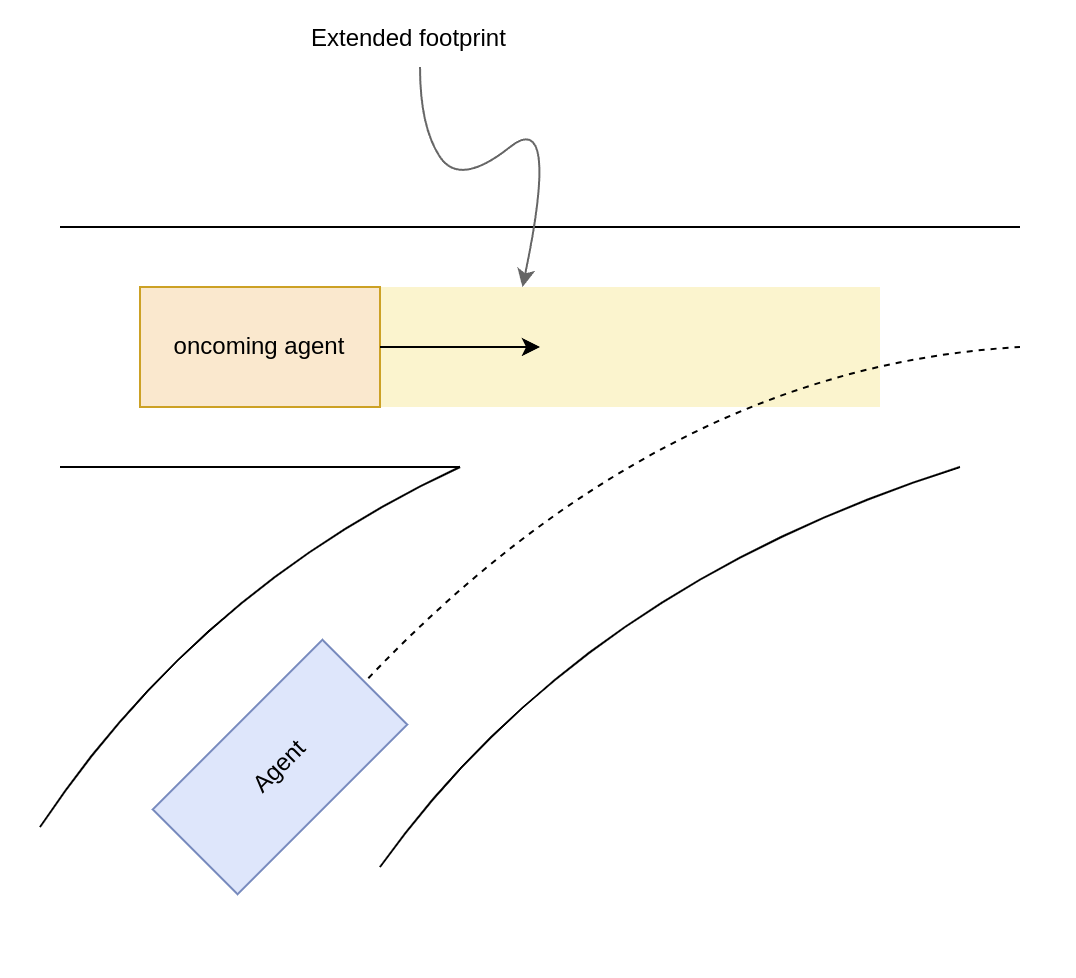}
    \caption{Footprint projection of other objects in the scene.}
    \label{fig:idm_footprint_projection}
\end{figure}

Merging situations can be tricky because IDM does not inherently account for multi-lane interactions. One way to simulate this is to extend all other objects' footprints in the scene proportional to their speed. Fig. \ref{fig:idm_footprint_projection} shows an example of a situation where an extended footprint can help in lane-merge situations. The lead agent search will identify the oncoming agent. The agent will know to slow down sooner, giving way to the oncoming objects.

The final IDM algorithm looks as such:
\begin{enumerate}
    \item Breadth-first search path planning for a set distance.
    \item Identifying the closest leading object.
    \item Perform a one-step forward euler numerical integration on the IDM differential equations.
    \item Propagate the agent along the planned path according to the solution of step 3.
    \item Repeat for all agents in the scene.
    \item Repeat for all simulation propagation steps.
\end{enumerate}

The attempt to re-purpose prevailing motion forecasting models such as LaneGCN for traffic simulation proved less trivial than initially thought. 
The model independently predicts agents, resulting in a lack of scene cohesion, for example, predicting colliding trajectories. The model is also susceptible to distribution shift issues that arise in closed-loop simulation. When applied to traffic simulation, the model induces unrealistic driving scenarios \cite{igl2022symphony}. Furthermore, the number of simulated agents had to be limited to maintain acceptable simulation runtime. It can be concluded that motion forecasting alone cannot achieve realistic, scene-coherent traffic simulation.

\subsection{Evaluation}
\label{sec:appendix:simulation:evaluation}

\textbf{Metrics.}
As explained in the main paper, we design a set of metrics along with a scoring function to compare the performance of planners. 
The previously mentioned open-loop metrics are described in detail in Tab.~\ref{tab:olm}.
For each metric, we evaluate an aggregated error/MR considering different time horizons within the planning horizon (i.e., $H=3,5,8s$) and with the same sampling frequency of $1 Hz$. 
This method allows us to have a fair comparison across planners with different planning horizons or sampling rates. Additionally, by taking the mean across the selected horizons, the errors at the beginning of the horizon will have more impact on the averaged value. The "within bound" metric score used in the cost structure is found by comparing the average error/MR value to a maximum acceptable threshold ($8m$ for distance errors, $0.8rad$ for heading errors, and $0.3$ for MR).
It is $0$ if the average value is more than the threshold, and $1$ otherwise. 
For example, `MR within bound' from open-loop metrics, and `drivable area compliance' from the closed-loop metrics.
Multiplier metrics are assigned a score of 1 or 0, except for `no at-fault collision' which takes 0 (if there is an at-fault collision with a vehicle, bicycle, or pedestrian, or there are multiple at-fault collisions with objects), 0.5 (if there's an at-fault collision with a single object), and 1 (if there's no at-fault collision).
\begin{table}
    \centering
    \caption{Open-loop metrics and their definitions
    }

    \label{tab:olm}
    \begin{tabular}{  L{2cm} | L{6cm}  }
     \bottomrule
     Metric name & Metric definition\\
     \hline
     ADE Within Bound & At each sampled time, ADE is defined as the average of pointwise L2 distances between the planner trajectory (x-y) and expert trajectory, up to the selected comparison horizon in the future. 
     \\
     \hline
     FDE Within Bound & At each sampled time, FDE is defined as the L2 distance between the planner trajectory (x-y) and expert trajectory at the final time available in the sampled trajectories.
     \\
     \hline
     AHE Within Bound & At each sampled time, AHE is defined as the average of absolute differences between the planner trajectory heading and expert trajectory heading up to the selected comparison horizon in the future. 
     \\
     \hline
     FHE Within Bound  & At each sampled time, FHE is defined as the absolute differences between  the planner trajectory heading and expert trajectory heading at the final time available in the sampled trajectories.
     \\
     \hline
     Miss Rate Within Bound & At each sampled time, if the maximum of the pointwise L2 distances between the planner trajectory and expert trajectory up to the selected comparison horizon in the future is greater than its corresponding maximum displacement threshold (6, 8 and 16m for horizons 3, 5 and 8s, respectively), we consider the planner trajectory at that time as a miss. Miss rate is the ratio of sampled times where the trajectory was marked as a miss over the number of the sampled times.
     \\
     \hline
    \toprule
    \end{tabular}
\end{table}

In the following, we include additional information about closed-loop metrics mentioned in the main paper:
\begin{itemize}
\item When identifying at-fault collisions, we only penalize the planner when the ego vehicle could be responsible for the collision, which includes collisions with stopped agents, collisions with agents in front of ego and collisions with agents in adjacent lanes while making a lane change.
On the other hand, the ego is not penalized for rear-end collisions or other agents colliding with the ego when it is stopped.
To further emphasize the importance of different agent types, at-fault collisions are grouped into vulnerable road users (including pedestrians and bicyclists), vehicles and objects (traffic cones, barriers and generic objects). 
\item For drivable area violation, we measure the maximum distance of the corners of the ego bounding box from the nearest drivable area. 
\item For driving direction, the movement of the ego during a 1s time horizon is calculated along the driving direction of its lane.
\item Speed limit violation is defined based on the magnitude and duration of the violation.
\item Time to collision is defined as the time required for ego and another object to collide if they continue at their present speed and heading.
We only compute time to collision for objects in front of the ego, cross-traffic objects and lateral objects on the sides, when the ego is making a lane change or is in the intersection. 
\item Rider comfort is measured based on jerk, acceleration and steering rate which are compared to those observed in human driving.
\item Progress of the planner trajectory towards the goal is evaluated by comparing its progress along expert’s route in the same scenario.
The metric quantifies the progress as the ratio of overall ego progress to the overall expert’s progress during the scenario.
\end{itemize}
\textbf{Final score structure.}
The final score of a planner is computed by averaging its scores across all scenarios as defined in the main paper.
What follows helps explain the equation:
For open-loop planners, the driven trajectory in a scenario is assigned a zero score if the miss rate is above the selected threshold (0.3), otherwise, a weighted average of other metrics’ scores is used as the score. All weights were tuned with the objective to maximize the overall performance of the planner against human-driven future trajectories across multiple scenarios. The weights can be found in the main paper.
For closed-loop planners, the driven trajectory in a scenario is assigned: 
\begin{itemize} 
\item A zero score if 1) there is an at-fault collision with a vehicle or a VRU, or 2) there are multiple at-fault collisions with objects (e.g. a cone), or 3) there is a drivable area violation, or 4) ego drives into oncoming traffic more than $6m$ (driving distance), or 5) ego progress towards the destination is smaller than a threshold.
\item The weighted average of other metrics’ scores is multiplied with $0.5$ if there is one at-fault collision with an object (e.g. a cone), or if ego drives into oncoming traffic more than $2m$, but less than $6m$.
\item A weighted average of other metrics’ scores, otherwise.
\end{itemize}
Closed-loop metric scores are summarized in Tab.~\ref{tab:metric2}.
$f_{speed-limit}$ is a function that returns 1 if there are no speed limit violations and approaches 0 as the violation increases.
Furthermore, the comfort metric accounts for jerk amplitude, lateral and longitudinal acceleration, and jerk, and yaw rate and acceleration.
It is assigned a zero score if one of the comfort bounds is violated.
Our scoring heuristic described above is an initial proposal that accounts for the natural importance of each metric and is hand-tuned for our dataset and simulation framework.

\begin{table}
    \centering
    \caption{Closed-loop metrics, scores and weights}
    \label{tab:metric2}
    \vspace{-3mm}
    \begin{tabular}{ l|c }
     \bottomrule
     Metric name & Metric score \\
     \hline
     No at-fault Collisions & 0, 0.5 or 1 \\
      Drivable Area Compliance& boolean \\
      Driving Direction Compliance&  0, 0.5 or 1 \\
    Making progress & boolean \\
      Speed limit compliance& $f_{speed-limit}$ \\
        Time to Collision within bound &  boolean \\
      Progress along route & $\frac{\text{ego progress}}{\text{expert progress}}$ \\
      Comfort & boolean \\
    \toprule
    \end{tabular}
\end{table}

\section{Experiments}

\subsection{Planning experiments}
\label{sec:appendix:experiments:planning}
Detailed scenario-stratified metrics for all four planner baselines (rule-based and learned) across the three challenges (open-loop, closed-loop non-reactive, and closed-loop reactive) can be found in Tab.~\ref{tab:planning_experiment_metrics}. The metric breakdown corroborates the statement in the main paper that metrics reward conservative driving. Planners that do not collide and stay within the drivable area may score better than planners that attempt to mimic human drivers. Hence, finer grain and scenario-based metrics are required to distinguish between simple rules-abiding driving from desirable human-like driving behaviors.

\begin{table*}[t]
    \caption{The metrics breakdown by scenario types across the three challenges for all planners: SimplePlanner (SP), IDMPlanner (IDM), Raster Planner (RP), and UrbanDriver (UD)
    }
\begin{tabularx}{\textwidth}{l | lrrrrllrrrrllrrrr}
\hline
\multirow{2}{*}{Scenario type} &
   &
  \multicolumn{4}{c}{Open-loop} &
   &
   &
  \multicolumn{4}{c}{Closed-loop Non-reactive} &
   &
   &
  \multicolumn{4}{c}{Closed-loop Reactive} \\
 &
   &
  \multicolumn{1}{c}{SP} &
  \multicolumn{1}{c}{IDM} &
  \multicolumn{1}{c}{RP} &
  \multicolumn{1}{c}{UD} &
   &
   &
  \multicolumn{1}{c}{SP} &
  \multicolumn{1}{c}{IDM} &
  \multicolumn{1}{c}{RP} &
  \multicolumn{1}{c}{UD} &
   &
   &
  \multicolumn{1}{c}{SP} &
  \multicolumn{1}{c}{IDM} &
  \multicolumn{1}{c}{RP} &
  \multicolumn{1}{c}{UD} \\ \hline
all                             &  & 0.22 & 0.30 & 0.52 & 0.90 &  &  & 0.32 & 0.73 & 0.47 & 0.68 &  &  & 0.37 & 0.76 & 0.46 & 0.67 \\
behind long vehicle             &  & 0.74 & 0.54 & 0.79 & 0.95 &  &  & 0.17 & 0.99 & 0.79 & 0.99 &  &  & 0.44 & 1.00 & 0.90 & 0.98 \\
changing lane                   &  & 0.16 & 0.29 & 0.41 & 0.89 &  &  & 0.35 & 0.61 & 0.43 & 0.81 &  &  & 0.41 & 0.59 & 0.42 & 0.74 \\
following lane with lead        &  & 0.15 & 0.05 & 0.30 & 0.89 &  &  & 0.40 & 0.77 & 0.48 & 0.77 &  &  & 0.47 & 0.91 & 0.26 & 0.82 \\
high lateral acceleration       &  & 0.07 & 0.32 & 0.41 & 0.86 &  &  & 0.13 & 0.84 & 0.24 & 0.58 &  &  & 0.14 & 0.85 & 0.22 & 0.54 \\
high magnitude speed            &  & 0.02 & 0.15 & 0.50 & 0.90 &  &  & 0.64 & 0.77 & 0.57 & 0.89 &  &  & 0.71 & 0.89 & 0.47 & 0.90 \\
low magnitude speed             &  & 0.49 & 0.45 & 0.66 & 0.92 &  &  & 0.29 & 0.83 & 0.57 & 0.68 &  &  & 0.38 & 0.80 & 0.61 & 0.79 \\
near multiple vehicles          &  & 0.20 & 0.26 & 0.37 & 0.93 &  &  & 0.45 & 0.82 & 0.61 & 0.89 &  &  & 0.52 & 0.84 & 0.50 & 0.80 \\
starting left turn              &  & 0.00 & 0.13 & 0.39 & 0.85 &  &  & 0.21 & 0.56 & 0.07 & 0.48 &  &  & 0.21 & 0.63 & 0.17 & 0.51 \\
starting right turn             &  & 0.00 & 0.18 & 0.41 & 0.87 &  &  & 0.09 & 0.33 & 0.20 & 0.14 &  &  & 0.09 & 0.40 & 0.27 & 0.12 \\
intersection traversal          &  & 0.02 & 0.21 & 0.47 & 0.91 &  &  & 0.55 & 0.81 & 0.53 & 0.78 &  &  & 0.57 & 0.78 & 0.44 & 0.77 \\
stationary in traffic           &  & 0.73 & 0.70 & 0.87 & 0.97 &  &  & 0.71 & 0.97 & 0.86 & 0.94 &  &  & 0.73 & 0.97 & 0.87 & 0.93 \\
stopping with lead              &  & 0.54 & 0.54 & 0.79 & 0.95 &  &  & 0.09 & 0.95 & 0.77 & 0.91 &  &  & 0.25 & 0.95 & 0.72 & 0.84 \\
traversing pickup dropoff       &  & 0.18 & 0.26 & 0.45 & 0.87 &  &  & 0.17 & 0.68 & 0.35 & 0.49 &  &  & 0.17 & 0.71 & 0.37 & 0.48 \\
waiting for pedestrian to cross &  & 0.13 & 0.17 & 0.51 & 0.81 &  &  & 0.09 & 0.30 & 0.29 & 0.22 &  &  & 0.13 & 0.38 & 0.37 & 0.24 \\ \hline
\end{tabularx}
\label{tab:planning_experiment_metrics}
\end{table*}

\subsection{nuPlan Challenge}
\begin{figure}
    \centering
    \includegraphics[width=0.4\textwidth]{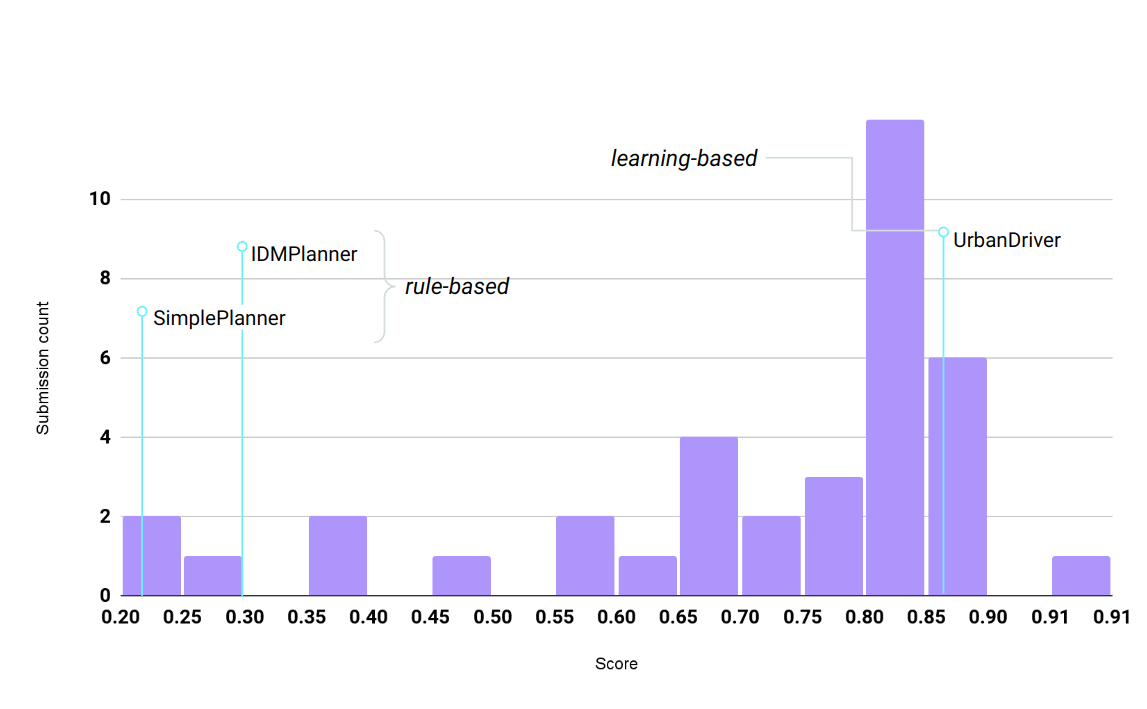}
    \caption{nuPlan challenge open-loop score distribution.
    }
    \label{fig:open-loop-histogram}
\end{figure}

\begin{figure}
    \centering
    \includegraphics[width=0.4\textwidth]{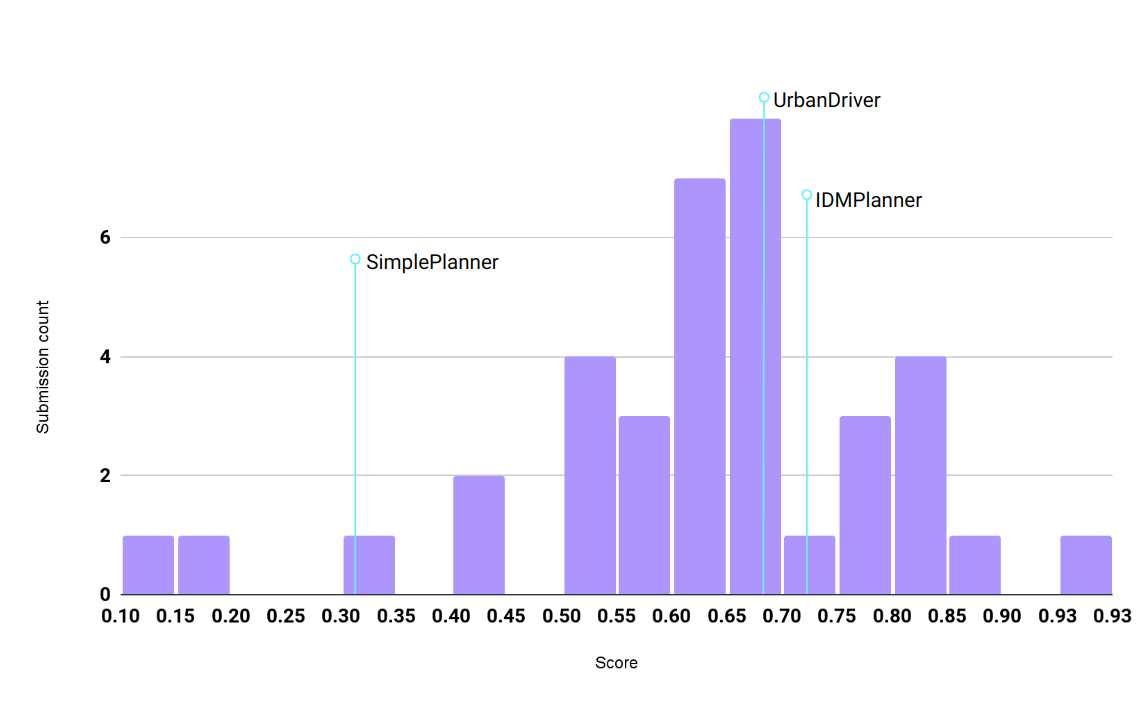}
    \caption{nuPlan challenge closed-loop non-reactive score distribution.
    }
    \label{fig:closed-loop-nonreactive-histogram}
\end{figure}

\begin{figure}
    \centering
    \includegraphics[width=0.4\textwidth]{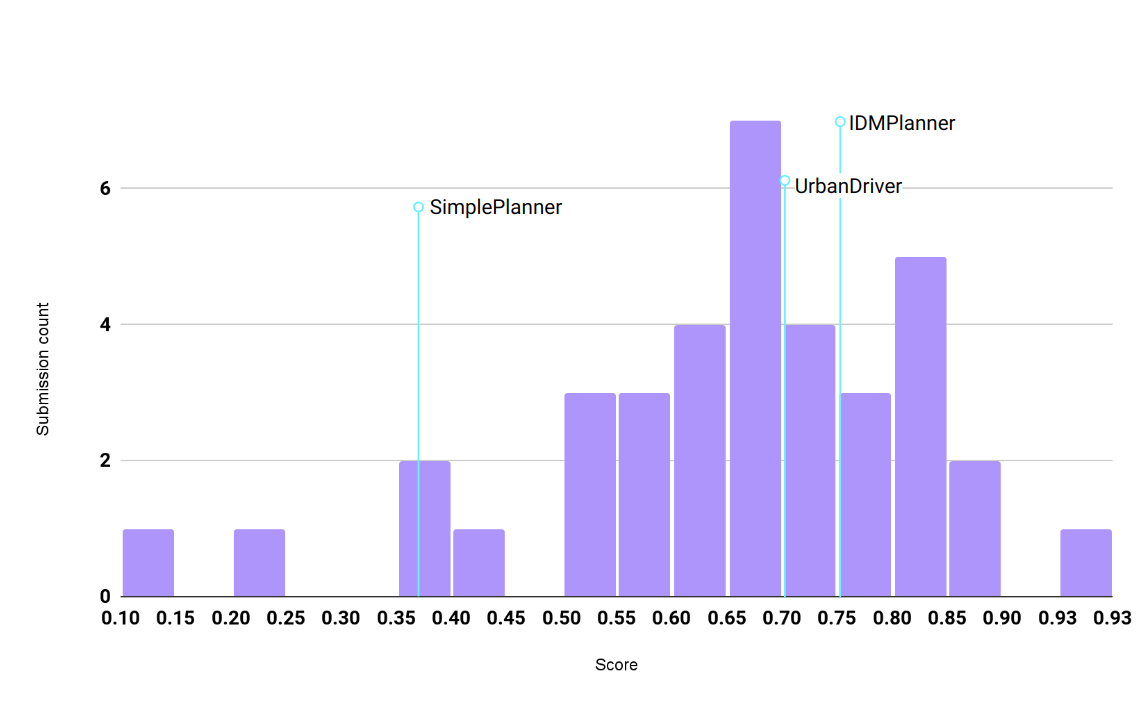}
    \caption{nuPlan challenge closed-loop reactive score distribution.
    }
    \label{fig:closed-loop-reactive-histogram}
\end{figure}

The distribution of planners' scores across the nuPlan challenge can be seen in Fig. \ref{fig:open-loop-histogram} for open-loop, Fig. \ref{fig:closed-loop-nonreactive-histogram} for closed-loop non-reactive and Fig. \ref{fig:closed-loop-reactive-histogram} for closed-loop reactive. Most submitted planners were able to score well in the open-loop challenge. The most common scores lie between 0.8 - 0.85. The scores significantly dropped for closed-loop challenges. The most common scores lie between 0.65 - 0.7. A 0.15 drop from the open-loop challenge. This further supports the argument that most purely learned models fail to generalize to closed-loop scenarios. For an overview of the leaderboard, please refer to \footnote{\url{https://eval.ai/web/challenges/challenge-page/1856/leaderboard/4360}}.

\begin{table}[t]
    \caption{Open-loop Challenge}
    \label{tab:open_loop_challenge}
    \begin{tabular}{l|cccc}
    \hline
    Metric                                & CS Tu & AutoHorizon    & Pegasus        & AID            \\ \hline
    Overall score                         & 0.829 & 0.852          & \textbf{0.876} & 0.840          \\
    ADE within bound                      & 0.815 & 0.836          & \textbf{0.843} & 0.811          \\
    FDE within bound                      & 0.597 & 0.665          & \textbf{0.727} & 0.639          \\
    Miss rate within bound                & 0.962 & 0.960          & 0.960          & \textbf{0.966} \\
    AHE within bound                      & 0.944 & \textbf{0.958} & 0.947          & 0.937          \\
    FHE within bound                      & 0.921 & \textbf{0.938} & 0.935          & 0.925          \\ \hline
    \end{tabular}
\end{table}

\begin{table}
    \caption{Closed-loop Non-reactive Challenge}
    \label{tab:closed_loop_nonreactive_challenge}
    \begin{tabular}{l|cccc}
    \hline
    Metric                          & CS Tu          & AutoHorizon    & Pegasus        & AID            \\ \hline
    Overall score                   & \textbf{0.928} & 0.890          & 0.817          & 0.809          \\
    Making progress                 & \textbf{0.994} & 0.978          & 0.929          & 0.936          \\
    Drivable area compliance        & \textbf{1.000} & 0.990          & 0.948          & 0.962          \\
    Driving direction compliance    & \textbf{1.000} & 0.988          & 0.955          & 0.992 \\
    Comfort                         & 0.919          & \textbf{0.990} & 0.927          & 0.940          \\
    No at-fault collisions          & \textbf{0.988} & 0.963          & 0.926          & 0.939          \\
    TTC within bound                & \textbf{0.925} & 0.905          & 0.879          & 0.883          \\
    Progress along expert route     & 0.914          & \textbf{0.915} & 0.793          & 0.841          \\
    Speed limit compliance          & \textbf{0.997} & 0.960          & 0.934          & 0.974          \\ \hline
    \end{tabular}
\end{table}

\begin{table}
    \caption{Closed-loop Reactive Challenge}
    \label{tab:closed_loop_reactive_challenge}
    \begin{tabular}{l|cccc}
    \hline
    Metric                          & CS Tu          & AutoHorizon    & Pegasus        & AID            \\ \hline
    Overall score                   & \textbf{0.929} & 0.881          & 0.851          & 0.838          \\
    Making progress                 & \textbf{0.992} & 0.980          & 0.946          & 0.946          \\
    Drivable area compliance        & \textbf{1.000} & 0.992          & 0.952          & 0.966          \\
    Driving direction compliance    & \textbf{1.000} & 0.992          & 0.962          & 0.996 \\
    ego is comfortable              & 0.925          & \textbf{0.992} & 0.952          & 0.972          \\
    No at-fault collisions          & \textbf{0.993} & 0.965          & 0.946          & 0.969          \\
    TTC within bound                & \textbf{0.954} & 0.921          & 0.907          & 0.919          \\
    Progress along expert route     & \textbf{0.885} & 0.882          & 0.799          & 0.828          \\
    Speed limit compliance          & \textbf{0.997} & 0.964          & 0.940          & 0.981          \\ \hline
    \end{tabular}
\end{table}